\documentclass{fairmeta}
\usepackage{amsmath}
\usepackage{enumerate} 
\usepackage{bm}
\usepackage{floatflt}
\usepackage{algpseudocode}
\usepackage{amsfonts}
\usepackage{amsthm}
\usepackage{newtxtt}
\usepackage{colortbl} 
\usepackage{cleveref}
\usepackage{diagbox} 
\usepackage[utf8]{inputenc}
\usepackage{textgreek}
\usepackage{colortbl}
\usepackage{nicematrix}
\usepackage{makecell}
\usepackage{float}
\usepackage{arydshln}
\usepackage[frozencache,cachedir=.]{minted}
\usepackage[justification=centering]{caption}
\usepackage{subcaption}
\usepackage{tcolorbox}
\usepackage{amssymb}
\usepackage{xspace}
\usepackage{wrapfig}
\usepackage{adjustbox}
\usepackage{tabularx}
\usepackage{booktabs}
\usepackage{mathtools}
\usepackage{wrapfig}
\usepackage{amssymb}
\usepackage{graphicx}

\usepackage[ruled,vlined]{algorithm2e} 

\usepackage{silence}
\makeatletter
\patchcmd{\wrong@fontshape}{\@gobbletwo}{}{}{}
\makeatother
\newtheorem{theorem}{Theorem}[]

\newtheorem{remark1}[theorem]{Remark}

\definecolor{upColor}{RGB}{17,138,21}
\definecolor{downColor}{RGB}{174,36,67}

\newcommand{\scr}[1]{{\scriptsize #1}}
\newcommand{\emphTab}[2]{{#1}\scr{(#2)}}
\newcommand{\up}[1]{\textcolor{upColor}{#1}}
\newcommand{\down}[1]{\textcolor{downColor}{#1}}

\title{
    \raisebox{-0.2em}{\includegraphics[height=1em]{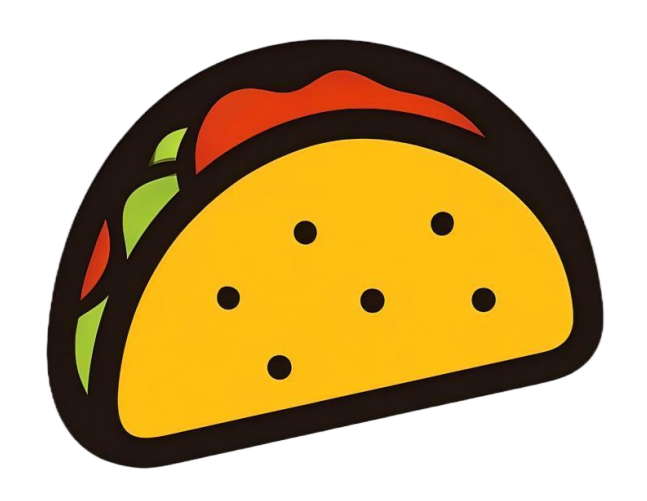}}
    Steering Vision-Language-Action Models as Anti-Exploration: A Test-Time Scaling Approach
}
\author[1,2,*]{\text{Siyuan Yang}}
\author[1,3,*,\ddagger]{\text{Yang Zhang}}
\author[4]{\text{Haoran He}}
\author[4]{\text{Ling Pan}}
\author[3]{\text{Xiu Li}}
\author[1,\dagger]{\text{Chenjia Bai}}
\author[1,\dagger]{\text{Xuelong Li}}
\affiliation[1]{Institute of Artificial Intelligence, China Telecom}
\affiliation[2]{University of Science and Technology of China}
\affiliation[3]{Tsinghua University}
\affiliation[4]{The Hong Kong University of Science and Technology}
\contribution{\textsuperscript{*}Equal Contributions$\quad$\textsuperscript{$\dagger$}Corresponding Authors$\quad$\textsuperscript{$\ddagger$}Project Leader}
\date{December 2, 2025}
\project{\url{https://vla-anti-exploration.github.io/}}
\code{\url{https://github.com/breez3young/TACO/}}
\metadata[Correspondence to]{Chenjia Bai (\email{baicj@chinatelecom.cn})}

\begin{document}

\abstract{

Vision-Language-Action (VLA) models, trained via flow-matching or diffusion objectives, excel at learning complex behaviors from large-scale, multi-modal datasets (e.g., human teleoperation, scripted policies).
However, since VLAs incorporate diverse data modes in the pre-training stage, and the finetuning dataset often contains demonstration data collected in a kinematically suboptimal or undesirable way, it exists redundant action modes that are irrelevant to the success action modes of the downstream task. Specifically, we observe a critical inference-time fragility among various sampled noises after supervised finetuning of pre-trained VLAs.
In this paper, we attribute this instability to the distribution shift between the VLA policy and the policy induced by stable success modes of the downstream task dataset.
Thus, we propose \textbf{TACO}, a test-time-scaling (TTS) framework that applies a lightweight pseudo-count estimator as a high-fidelity verifier of action chunks. The VLA models integrated with TACO can execute the actions with maximum pseudo-count from all sampled action chunks, thereby preventing distribution shifts while preserving the generalization ability of VLAs since the constraint is applied only during inference. Our method resembles the classical anti-exploration principle in offline reinforcement learning (RL), and being gradient-free, it incurs significant computational benefits compared to RL update, especially for flow or diffusion-based VLAs which are difficult to perform RL update due to denoising process. Extensive experiments across four simulation benchmarks (RoboTwin2.0, Robotwin, LIBERO, SimplerEnv) and a dual-arm platform demonstrate that our method significantly improves the inference stability and success rates in downstream-task adaptations. 
}


\maketitle

\section{Introduction}
\label{sec:intro}

Vision-Language-Action (VLA) models \citep{brohan2022rt1, zitkovich2023rt2, kim2024openvla, black2024pi0} have demonstrated remarkable performance and strong generalization capabilities in robotic manipulation tasks, benefiting from large-scale, multimodal datasets collected via human teleoperation or scripted policies. During pretraining, VLA models that employ flow-matching \citep{liu2023flow,lipman2023flow,lipman2024flowguide} or diffusion~\citep{sohl2015deep,ho2020ddpm,song2021ncsn}~-based objectives can effectively learn multimodal behavioral policies that span diverse manipulation modes, scenes, and tasks \citep{jang2021bc_z, openxembodiment, walke2023bridgedatav2, khazatsky2024droid}. Subsequently, the pretrained VLA model is adapted to downstream tasks through supervised fine-tuning (SFT) on small-scale, task-specific datasets. Leveraging a pretrained VLA for downstream adaptation typically yields substantially better generalization ability than training a policy from scratch, as the model benefits from rich pretrained knowledge—including visual perception, language understanding, embodiment priors, and awareness of object types and spatial configurations \citep{driess2023palm, beyer2024paligemma, wang2024qwen2vl}.

\begin{wrapfigure}{r}{0.5\linewidth}
    \centering
    \vspace{-0.5em}
    \includegraphics[width=1\linewidth]{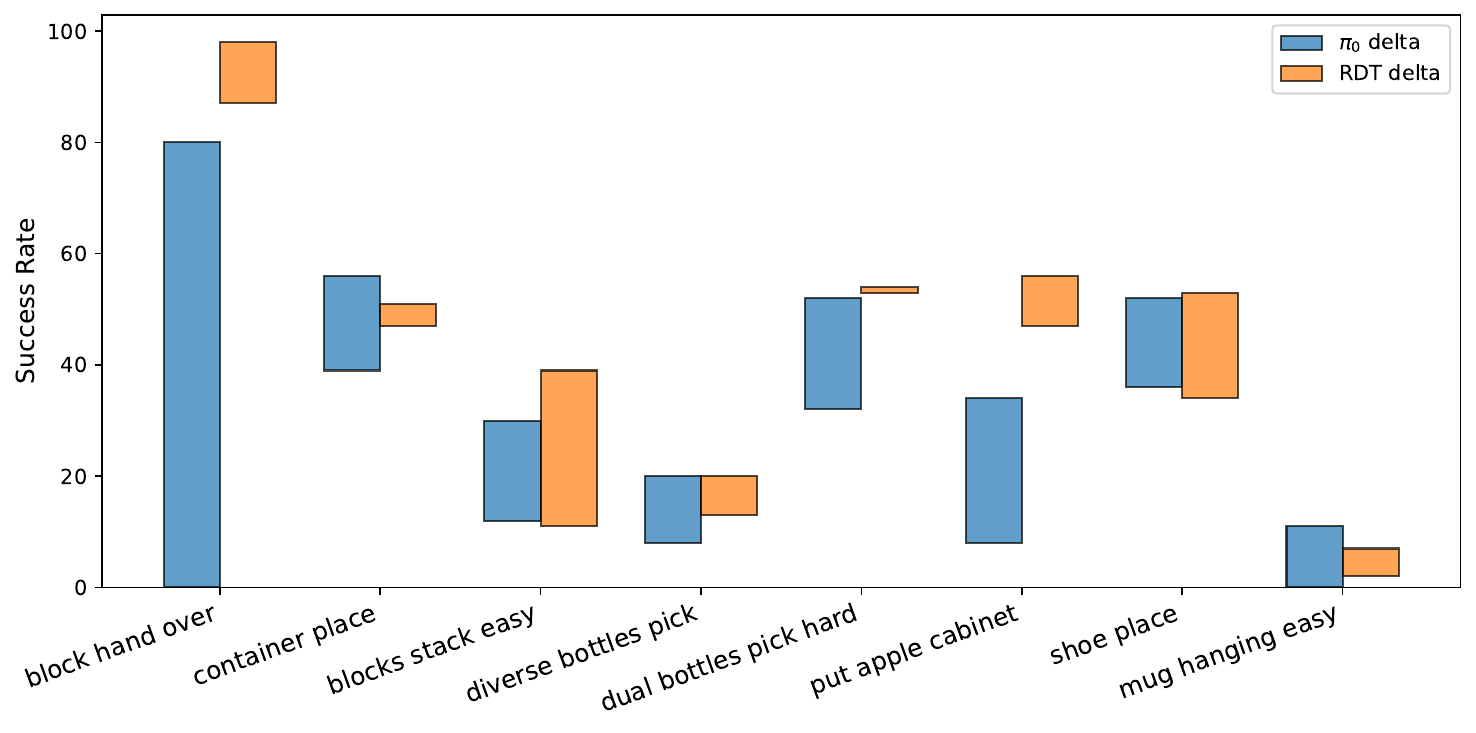}
    \vspace{-0.5em}
    \caption{\textbf{Illustrative Example:}
    We fix the initial noise for the denoising process of $\pi_{0}$ to ${\rm noise}_{1}$ and evaluate the success rate. Then we fix the initial noise for the denoising process of $\pi_{0}$ to another noise value ${\rm noise}_{2}$ and evaluate the success rate under the same scenarios. We plot the two resulting success rates using a floating bar chart. The same procedure is applied to RDT.
    }
    \vspace{-0.5em}
    
    \label{fig:intro-example}
\end{wrapfigure}

Although VLA models exhibit strong average performance, we observe a critical fragility at inference time in representative flow-matching \citep{black2024pi0} and diffusion-based VLA \citep{liu2025rdt} architectures—even after SFT on task-specific datasets \citep{mu2025robotwin}. As illustrated in Figure~\ref{fig:intro-example}, we evaluate the same SFT-adapted VLA model using two fixed noise vectors during inference across different tasks and find that the resulting success rates vary drastically—from 0\% to 80\%—solely due to noise variation. 

We attribute this instability to the persistence of redundant action modes that are irrelevant to the success action modes of the downstream task after SFT. Specifically, (\romannumeral1) during pretraining, VLA models absorb a broad spectrum of action modes from diverse data sources, making it difficult to rapidly narrow their output distribution to the narrow set of successful behaviors required by a specific downstream task. Consequently, the policy distribution after SFT still retains extraneous modes unrelated to task success. (\romannumeral2) SFT datasets themselves may exhibit multimodality, as they are often collected from multiple human teleoperators, scripted planners, or varying execution styles—some of which encode suboptimal or undesirable strategies. These redundant modes induce a significant distributional shift between the VLA policy and the ideal policy corresponding to the stable success mode of the downstream task. This shift becomes evident when sampling different noise vectors in flow-matching or diffusion-based frameworks, as the stochasticity of the sampled initial noise directly influences the resulting actions.

In this paper, we address the distribution shift problem from an anti-exploration perspective and propose \textbf{T}est-time \textbf{A}nti-exploration via pseudo-\textbf{CO}unts (\textbf{TACO}). The anti-exploration principle \citep{rezaeifar2022offline} originates from offline reinforcement learning (RL) \citep{levine2020offline_review}, where the goal is to prevent the policy from visiting states or actions that lie outside the support of the dataset. Analogously, during VLA inference, we aim to constrain the generated actions to lie within the support of the successful modes present in the SFT dataset—avoiding exploration of redundant or irrelevant action modes retained from pretraining or imperfect fine-tuning data. In TACO, we realize anti-exploration through test-time scaling (TTS) rather than policy optimization, as flow-matching or diffusion-based VLA models involve complex sampling dynamics that are not easily amenable to standard RL-style updates. Instead, TTS adjusts the action sampling process without modifying the VLA’s parameters, thus requiring no gradient computation.
We employ a classical Coin-Flipping Network (CFN) \citep{lobel2023cfn} enhanced with an internal representation mechanism to estimate pseudo-counts for every observation-instruction-action-chunk pair in the SFT dataset.
Intuitively, an action chunk with a higher pseudo-count is more consistent with the frequently observed (i.e., successful) behaviors in the SFT data and is thus preferred.
During inference, TACO uses the CFN-derived pseudo-counts as a verifier to select the most reliable action chunk among multiple action chunk candidates, ultimately executing the one with the highest pseudo-count.
To mitigate the computational overhead of repeated forward passes at test-time, we further introduce a shared observation key-value cache, which dramatically reduces latency by reusing visual-language representations across samples.

Our contributions can be summarized as follows:
\begin{itemize}
\item We propose TACO, a test-time scaling framework for VLAs which retains the strong generalization capabilities of pretrained VLAs while effectively constraining outputs to the success modes of specific downstream tasks.
\item We introduce an efficient internal representation mechanism for pseudo-count estimation in CFN, enabling accurate measurement of distributional shift with minimal computational overhead.
\item We demonstrate that TACO significantly boosts the success rates of diverse VLA models across both simulation and real-world tasks without prolonged training time and can operate with low latency.
\end{itemize}

\section{Related Work}

\textbf{Vision-Language-Action Models.}
Vision-Language-Action (VLA) models have emerged as a dominant paradigm for general-purpose robotic manipulation, integrating the reasoning capabilities of pre-trained VLMs with robotic control.
While early works like RT-1 and Octo \citep{brohan2022rt1, team2024octo} trained policies from scratch, later approaches such as RT-2 and OpenVLA \citep{zitkovich2023rt2, kim2024openvla} successfully leveraged pre-trained backbones (e.g., PaLI-X, Prismatic)~\citep{chen2023pali, anschutz2023prismatic} via action discretization.
To better capture the multimodality and precision of action distribution in large-scale cross-embodiment action-labeled datasets \citep{openxembodiment, wu2024robomind, bu2025agibot}, recent approaches have shifted towards generative policies. Methods like RDT-1B, DexVLA, DexGraspVLA, Dita \citep{liu2025rdt, wen2025dexvla, zhong2025dexgraspvla, hou2025dita} introduced diffusion \citep{sohl2015deep, ho2020ddpm, song2021ncsn} architectures, and the latest generation of models—including $\pi_0$, GR00T N1, and SmolVLA \citep{black2024pi0, bjorck2025gr00tn1, shukor2025smolvla}—leverage Mixture-of-Transformers architecture and employ flow matching \citep{liu2023flow,lipman2023flow, lipman2024flowguide} or diffusion experts to synthesize continuous action sequences.
Addressing the distribution shift in these score-based generative frameworks, recent work has also explored latent alignment. For instance, ATE~\citep{zhang2025ate} introduces a unified latent space to align disparate action distributions and employs classifier guidance during training time, to steer denoising-based VLAs for efficient adaptation.
Despite this rapid architectural progress, a critical and often overlooked issue is the inference-time instability of these generative models. Our work is the first to systematically diagnose this instability as an out-of-support problem and propose a test-time correction mechanism based on anti-exploration.

\noindent\textbf{Anti-exploration \& Offline RL.}
In online RL \citep{Sutton1999PolicyGM}, the agent actively interacts with the environment to explore the state–action space \citep{Yang2021ExplorationID}. In contrast, offline RL learns a policy exclusively from a fixed, previously collected dataset, without further environmental interaction \citep{Levine2020OfflineRL,BridgingOR}. This setting poses significant challenges—even for off-policy methods—due to distributional shift between the behavior policy that generated the data and the target policy being learned \citep{Wu2019BehaviorRO}. Specifically, the offline dataset reflects the visitation distribution of the behavior policy, which may differ substantially from that of the learned policy \citep{Agarwal2019AnOP}. This discrepancy can cause severe extrapolation errors when estimating values for out-of-distribution (OOD) state–action pairs \citep{kumar2020cql,bai2022pessimistic}. A similar distribution shift arises during the SFT stage of VLAs, where the VLA model encode a broad and redundant action modes, whereas successful behaviors in downstream tasks typically occupy a much narrower region. Consequently, when sampling actions in flow or diffusion-based process, the policy may generate suboptimal behaviors \citep{lipman2023flow,ho2020ddpm}. In TACO, we employ an anti-exploration strategy \citep{rezaeifar2022offline,nikulin2023anti} that imposes policy constraints to restrict the policy to stay close to the support of the success data distribution. Unlike conventional offline RL approaches that typically rely on uncertainty-aware or conservative penalties \citep{TD3BC,Kidambi2020MOReLM,an2021uncertaintybased}, our method adopts a simple count-based mechanism \citep{ostrovski2017count,NIPS2016_afda3322,lobel2023cfn}, which is well-suited to the structure of VLA models and can operate without of value estimation \citep{kostrikov2022iql}. Notably, count-based regularization has precedent in both classical RL and recent RLHF algorithms.

\noindent\textbf{{Test Time Scaling in VLA.}}
Performance improvements at test time by extending inference have been widely validated in LLMs \citep{codemonkeys, AlphaMath, brown2024large}. 
In VLA research, many works achieve performance gains through internal `thinking' mechanisms, but most of them require datasets with annotated reasoning traces \citep{Zhao2025CoTVLAVC}. 
In contrast, test-time scaling introduces an additional scoring module to select better candidates, enabling a generate-then-justify mechanism to further boost the performance of the policy without modifying the network weights, as demonstrated by methods that utilized the advantage function \citep{zhang2025read} and state-action value function \citep{nakamoto2024vgps}, and ohther variants like Verifier-free \citep{Jang2025VerifierfreeTS} that considered the likelihood.
Recent work \citep{du2025dynaguide} also explored leveraging the distance to the goal within the representation of a learned world model \citep{hafner2019dreamv1, hafner2020dreamerv2, zhang2025marie, zhang2025dima} as a metric to steer the base policy.
Repeated sampling combined with a verifier has also been shown effective in several works \citep{song2021ncsn, RoboMonkey}; however, these approaches typically involve complex RL training or rely on verifiers with a large number of parameters. To the best of our knowledge, no prior work has leveraged the internal features of a VLA to build a lightweight and efficient verifier that fully exploits the model's understanding capabilities, without modifying the backbone network.
\section{Preliminaries}\label{sec:preliminaries}
\textbf{Problem Statement.} We consider a language-conditioned robotic manipulation task with vision input under the imitation learning setting. We assume access to a pre-collected dataset $\mathcal{D}_\text{sft} = \{ \tau_1, \tau_2, \ldots, \tau_n \}$ comprising $n$ demonstrations from one or more target downstream tasks.
Each demonstration $\tau = (\mathbf{o}_{1:T}, l, \mathbf{a}_{1:T})$ contains a natural language instruction $l$ describing the {target task}, a sequence of observations $\mathbf{o}_{1:T}$, where each observation $\mathbf{o}_t = (\mathbf{I}_t^1, \ldots, \mathbf{I}_t^n, \mathbf{q}_t)$ consists of multiple RGB images $(\mathbf{I}_t^1, \ldots, \mathbf{I}_t^n)$ and the robot’s proprioceptive state $\mathbf{q}_t \in \mathbb{R}^m$, and a sequence of actions $\mathbf{a}_{1:T}$ where $\mathbf{a}_t \in \mathbb{R}^n$.
In practice, expert demonstrations in $\mathcal{D}_\text{sft}$ are scarce due to the high cost of obtaining them.
Thus, they typically cover a limited fraction of the whole observation and action spaces, which in turn makes the distribution shift issue potentially more pronounced.

\noindent\textbf{Coin Flipping Network.}
Coin Flipping Network (CFN) \citep{lobel2023cfn} is a neural estimator which uses the sampling distribution of Rademacher trials (or \emph{coin flips}) made every time a state is encountered to derive the state visitation counts.
Given a state \( s \), each occurrence of \( s \) is paired with a random binary vector 
\(
\mathbf{c}_i \sim \{-1, 1\}^d,
\)
forming a dataset \( \mathcal{D}_{\text{cfn}} = \{(s_i, \mathbf{c}_i)\} \).
The CFN $f_\phi$ parameterized by $\phi$ is learned by solving the following simple regression problem:
\begin{align}\label{equ:cfn_loss}
    \min\nolimits_\phi \mathbb{E}_{(s_i, \mathbf{c}_i) \sim \mathcal{D}_{\text{cfn}}}
\big[\|f_\phi(s_i) - \mathbf{c}_i\|^2\big].
\end{align}
For a state \( s \) appearing \( m \) times in \( \mathcal{D}_{\text{cfn}} \), the optimal solution satisfies
\(
f_\phi^*(s) = \frac{1}{m}\sum_{i=1}^m \mathbf{c}_i. 
\)
If each element $c_{ij}$ in \( \mathbf{c}_i \) follows a fair coin flipping distribution,
\begin{align}
     \frac{1}{d}\|f_\phi(s)\|^2 = \frac{1}{d}\sum\nolimits_{j=1}^{d}\mathbb{E}\big[ ( \sum\nolimits_{i=1}^{m}\frac{c_{ij}}{m} ) \big] = \frac{1}{m}.
\end{align}
Thus, we can simply approximate the inverse of the state visitation count given by $\| f_{\phi}(s) \|^2 /d \approx 1/ N(s)$. 
In this work, the CFN would be utilized for \emph{anti-exploration}.

\section{Method}
\begin{figure*}[tbp]
    \centering
    \resizebox{0.95\textwidth}{!}{%
        \includegraphics{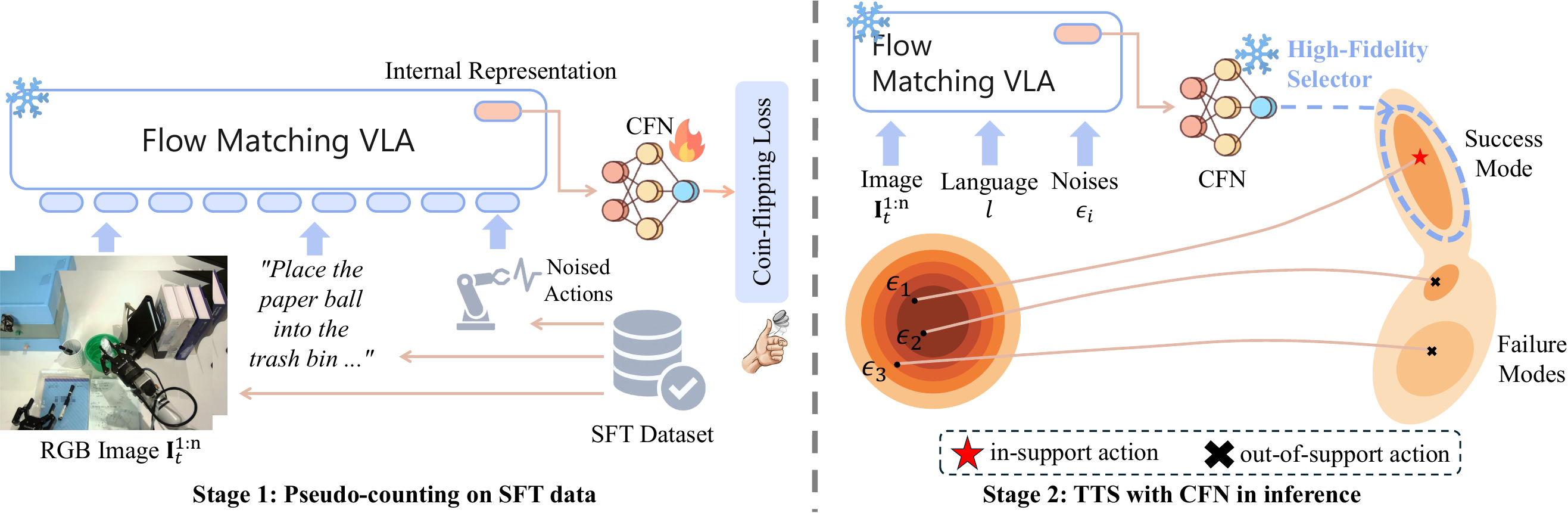}
    }
    \caption{\textbf{Overview of TACO.} 
    In the training stage (Stage 1), we sample data from the SFT dataset, add a certain amount of noise to the expert actions, and feed them into the VLA to denoise the actions while extracting internal representations. 
    These representations are then used to train the CFN. 
    During inference (Stage 2), the VLA generates multiple candidate actions along with their corresponding internal representations, and the CFN serves as a selector to select the action with the highest count for execution.
    }
    \label{fig:method_overview}
\end{figure*}

We now present TACO, a framework that treats the pseudo count estimator as an off-the-shelf verifier to scale test-time compute for VLAs, realizing \emph{Anti-Exploration} principle.
Note that TACO can be directly integrated into VLAs that either use discrete token auto-regression (e.g., OpenVLA \citep{kim2024openvla}) or a score-based generative formulation (e.g., RDT \citep{liu2025rdt}, $\pi_0$ \citep{black2024pi0}, $\pi_{0.5}$ \citep{black2025pi05}) to model the action distribution.
We first elucidate how we draw inspiration from \emph{Anti-Exploration} in offline RL to mitigate the distribution shift between the fine-tuned VLA and the desired success mode in the provided dataset of the downstream task (\S\ref{sec_methods:motivation}).
Then, we describe in detail how we operationalize the principle of \emph{Anti-Exploration} (\S\ref{sec_methods:implementation}) and incorporate it to drive VLAs from being out-of-support during inference time (\S\ref{sec_methods:tt_sampling}).
An overview of TACO is shown in Figure~\ref{fig:method_overview}.

\subsection{Inference Instability as an OOD Problem}\label{sec_methods:motivation}
We begin by formally identifying the source of inference-time instability (\S\ref{sec:intro}). VLAs trained via SFT \citep{black2024pi0, liu2025rdt} are compelled to approximate the entire dataset distribution $p_\mathcal{D}(a|s)$. This dataset is often an imperfect mixture $p_\mathcal{D} = w_1 \pi^* + \sum_{k>1} w_k p_k$, where $\pi^*$ is the desired success mode and $p_{k>1}$ represent sub-optimal or unexpected modes. The resulting policy, $\pi_\theta \approx p_\mathcal{D}$, is thus inherently multimodal. Instability arises when $\pi_\theta$ generates an action $\mathbf{a}_j \sim p_{j>1}(\mathbf{a}|\mathbf{o}, l)$ from an undesired mode. However, $\pi_\theta$, trained via offline imitation learning (IL), is mode-agnostic and cannot distinguish successful from unsuccessful modes.

This paradigm mirrors the core challenge in Offline RL \citep{levine2020offline_review}: mitigating {extrapolation error} \citep{fujimoto2019off, kumar2019stabilizing, kumar2020cql, kostrikov2022iql}. Generating an action from an undesired mode $p_{j>1}$ is analogous to an Offline RL agent sampling an action that is {out-of-support} relative to the target success mode $\pi^*$.
A principled solution in Offline RL is \emph{anti-exploration} \citep{rezaeifar2022offline}, which optimizes a penalized value function to constrain the policy to stay within the support of the dataset. This provides our starting objective:
\begin{align}
    \mathbf{a}^* = \arg\max\nolimits_\mathbf{a}[Q(s, a) - b(s, a)]
\end{align}
where $Q(s, a)$ is the action-value function and $b(s, a)$ is an anti-exploration penalty (or "bonus") that assigns high cost to out-of-support state-action pairs.

To relate this objective to our test-time inference framework, we first introduce a ideally simplified setting: we model the task as a contextual bandit where executing an action chunk $\mathbf{a}_{1:H}$ given $(\mathbf{o}, l)$ yields a binary success ($r=1$) or failure ($r=0$) reward.
Under this formulation, the Q-function becomes equivalent to the success probability: $Q(\mathbf{o}, l, \mathbf{a}_{1:H}) = P(r = 1 | \mathbf{o}, l, \mathbf{a}_{1:H})$. The anti-exploration objective thus translates to,
{\small
\begin{align}
\label{equ:anti_exploration_offline_rl}
    \mathbf{a}_{1:H}^* = \arg\max\nolimits_{\mathbf{a}_{1:H}}[P(r = 1 | \mathbf{o}, l, \mathbf{a}_{1:H}) - b(\mathbf{o}, l, \mathbf{a}_{1:H})].
\end{align}}
To solve Eq.~\eqref{equ:anti_exploration_offline_rl}, we introduce our core assumption: the downstream dataset $\mathcal{D}_\text{sft}$ for VLA fine-tuning is demonstrative. That is, we assume the density of demonstrations correlates with task success; high-density modes correspond to an ideal policy $\pi^*$ and low-density modes correspond to $p_k$. Formally,
\begin{align}
\label{equ:assumption_p_proportional_to_density}
     P(r = 1 | \mathbf{o}, l, \mathbf{a}_{1:H}) \propto p_{\mathcal{D}_\text{sft}}(\mathbf{o}, l, \mathbf{a}_{1:H}).
\end{align}
Let the anti-exploration penalty be a count-based bonus $b(\mathbf{o}, l, \mathbf{a}_{1:H}) = 1 / {N}_{\mathcal{D}_\text{sft}}(\mathbf{o}, l, \mathbf{a}_{1:H})$ where ${N}_{\mathcal{D}_\text{sft}}(\mathbf{o}, l, \mathbf{a}_{1:H})$ is the corresponding visit count in the dataset $\mathcal{D}_\text{sft}$. 

This instantiation formally simplifies the principled Offline RL objective in Eq.~\eqref{equ:anti_exploration_offline_rl} to a equivalent objective: finding the action chunk with the \textbf{maximum visitation count}, i.e., density:
\begin{align}\label{equ:final_form}
     \mathbf{a}_{1:H}^* = \arg\max\nolimits_{\mathbf{a}_{1:H}} [N_{\mathcal{D}_\text{sft}}(\mathbf{o}, l, \mathbf{a}_{1:H})].
\end{align}
This result provides a strong theoretical justification for our method. Our test-time scaling framework, which seeks the highest-density (i.e., in-support) action, is not merely a heuristic. It constitutes a principled and theoretically equivalent realization of the anti-exploration objective standard in Offline RL.

\subsection{Coupled Pseudo-Count Estimation for VLAs}\label{sec_methods:implementation}
To realize our anti-exploration principle (Eq.~\eqref{equ:final_form}), we must estimate the visit count $N_{\mathcal{D}_\text{sft}}(\mathbf{o}_t, l, \mathbf{a})$\footnote{In the following, we omit the subscript of action chunk for clarity.}. We can replace it with a pseudo-count $\hat{N}_{\mathcal{D}_\text{sft}}$ by training a pseudo-count estimator $f_\phi$ on $\mathcal{D}_\text{sft}$. Motivated by recent successes \citep{lobel2023cfn, bai2025copo} on applying pseudo-counts to LLM preference optimization, we instantiate this estimator as a Coin Flipping Network (CFN).

\textbf{The Coupled Estimator.}
A critical design choice is the input representation $z = \text{Enc}(\mathbf{o}, l, \mathbf{a})$. Instead of training a separate encoder, we hypothesize that the VLA $\pi_\theta$ itself provides the richest representations. We instantiate the CFN $f_\phi$ as a simple, lightweight MLP head that takes the VLA's internal representation $h_\theta$ as input (see Figure~\ref{fig:method_overview}). This {coupled estimator} design is highly efficient, leveraging the VLA’s computation, and benefits from its extensive pre-training.

\textbf{High-Fidelity Feature Search for Denoising-based VLAs.}
This coupled design introduces a critical challenge for diffusion-based or flow-based VLAs \citep{black2024pi0, liu2025rdt, black2025pi05}. These models are trained exclusively on {noised} actions $\{\mathbf{a}_\sigma\}$ and thus never see clean data actions $\{\mathbf{a}\}$ during training. Directly extracting $h_\theta$ by feeding $\mathbf{a} \in \mathcal{D}_\text{sft}$ may not lie on VLA's feature space, yielding uninformative representations.

Our goal is to find an in-distribution feature $h_\theta$ that best represents the clean action $a$. We propose a \textbf{High-Fidelity Feature Search} procedure. For each data point $(\mathbf{o}, l, \mathbf{a}) \in \mathcal{D}_\text{sft}$, we query the VLA $N$ times with different noise levels $\{\sigma_i\}_{i=1}^N$:
\begin{align}
    \{ (\mathbf{a}^{(i)}_{\text{pre}}, h^{(i)}_\theta) \}_{i=1}^N = \{ \text{VLA}(\mathbf{o}, l, \mathbf{a}_{\sigma_i}) \}_{i=1}^N
\end{align}
where $\mathbf{a}_{\sigma_i}$ is the corrupted version of the original action, $\mathbf{a}^{(i)}_{\text{pre}}$ is the VLA's resulting clean prediction, and $h^{(i)}_\theta$ is the corresponding internal representation.
We then select the feature $h^{(i^\star)}_\theta$ whose corresponding action prediction $\mathbf{a}^{(i^\star)}_{\text{pre}}$ is closest to the ground-truth $\mathbf{a}$:
\begin{align}
    i^\star = \arg\min\nolimits_{i \in \{1,\dots,N\}} \| \mathbf{a}^{(i)}_{\text{pre}} - \mathbf{a} \|_2.
\end{align}
The resulting feature $h^{(i^\star)}_\theta$ is now both \emph{in-distribution} for the VLA (as it was generated from a noised input $\mathbf{a}_{\sigma_{i^\star}}$) and \emph{high-fidelity} (as it is confirmed to represent $\mathbf{a}$).

\textbf{Estimator Training.}
This search procedure yields a high-fidelity feature set $\mathcal{D}_h = \{h^{(i^\star)}_\theta\}$ representing our original dataset $\mathcal{D}_\text{sft}$. We can now train the CFN head $f_\phi$ on top of any VLA's last hidden states.
Following \citep{lobel2023cfn}, the CFN $f_\phi$ is learned via optimizing Eq.~\eqref{equ:cfn_loss}.
After training the CFN, the pseudo-count of a feature $h_\theta$ is naturally given by
\begin{equation}\label{equ:pseudo_count}
\hat{N}_{\mathcal{D}_\text{sft}}(\mathbf{o},l,\hat{\mathbf{a}}) = \hat{N}_{\mathcal{D}_\text{sft}}(h_\theta) \propto \frac{1}{\| f_\phi(h_\theta) \|^2}.
\end{equation}
where $h_\theta$ is the feature extracted from a candidate action $\hat{\mathbf{a}}$ during inference.

As empirically validated in Figure~\ref{fig:count_dis}, the pseudo-counts derived from our coupled estimator exhibit a strong correlation between the L2 distance to the ground-truth action and the pseudo-count estimated by our CFN.
This confirms that our coupled CFN can effectively function as the verifier needed to penalize out-of-support actions.

\subsection{Test-Time Scaling as Anti-Exploration}\label{sec_methods:tt_sampling}
We now instantiate our anti-exploration framework through a two-stage \textbf{generate-then-verify} procedure during inference.
First, in the \emph{generation stage}, given an obsevration $\mathbf{o}_t$ and a language instruction $l$, we leverage the VLA $\pi_\theta$ which is already fine-tuned in the downstream dataset $\mathcal{D}_\text{sft}$, as a candidate generator.
For diffusion or flow-based models, this is achieved by sampling $M$ distinct initial noise vectors $\{\epsilon_i\}_{i=1}^M$ and performing the full denoising process in a batch-parallel way, yielding a batch of $M$ candidate action chunks $\{\hat{\mathbf{a}}^{(i)}_{t:t+H}\}_{i=1}^M$. Crucially, we simultaneously extract their corresponding internal representations $\{h_\theta^{(i)}\}_{i=1}^M$ during this generation pass.

Second, in the \emph{verification stage}, we deploy our trained CFN $f_\phi$ (in \S\ref{sec_methods:implementation}) as the verifier. The CFN scores each candidate $\hat{\mathbf{a}}^{(i)}_{t:t+H}$ based on its feature $h_\theta^{(i)}$, effectively estimating its pseudo-count $\hat{N}_{\mathcal{D}_\text{sft}} (\mathbf{o}_t,l,\hat{\mathbf{a}}^{(i)}_{t:t+H}) = 1 / \| f_\phi (h_\theta^{(i)}) \|^2$.
Finally, we select the single action $\hat{\mathbf{a}}^*_{t:t+H}$ that maximizes this pseudo-count,
\begin{align*}
\hat{\mathbf{a}}^*_{t:t+H} = \hat{\mathbf{a}}^{(i^*)}_{t:t+H}, \, \text{where} \,\,\, i^* = \arg\max\limits_{i \in \{1,\dots,M\}} \frac{1}{\| f_\phi (h_\theta^{(i)}) \|^2}.
\end{align*}
This procedure is a direct realization of our principled objective in Eq.~\eqref{equ:final_form}. Instead of taking a random sample from the VLA's potentially unstable multimodal distribution, we deterministically select the action that is most \textbf{in-support} (i.e., highest-count) w.r.t. the successful modes of $\mathcal{D}_\text{sft}$.

\textbf{Enabling Efficient Inference via KV Cache Optimization.}
Naively sampling $M$ candidates introduces a prohibitive $\mathcal{O}(M)$ overhead, rendering our method impractical. We solve this by observing that the expensive VLA computations (e.g., the transformer backbone) depend \emph{only} on the shared context tokens ($\mathbf{o}, l$). Therefore, we compute the Key (K) and Value (V) caches for this shared context just {once} and {re-use} them across all $M$ parallel action generation (e.g., denoising) processes. This optimization is a {key} of our method, making the marginal cost of additional candidates minimal. 
As shown in Figure~\ref{fig:real_exp} (left), when sampling with $M = 32$, our method reduces inference time by \textbf{73.2\%} compared to naive multi-batch parallel inference, making our high-performance anti-exploration sampling practical.
We provide an algorithmic description on how TACO does test-time scaling in \S\ref{app:algo}.

\section{Experiments}

Our experiments aim to verify whether the proposed Test-time Anti-exploration via pseudo-COunts  can effectively improve the accuracy of action prediction, thereby enhancing task success rates. We also seek to understand which design choices contribute to these improvements, and whether the additional computational cost is acceptable. 
Specifically, our experiments are designed to answer the following three research questions:

\textbf{Q1:} Can our framework improve performance across diverse environments, various VLA policies, and a wide range of tasks?

\textbf{Q2:} What is the additional time cost introduced by our test-time scaling method? Can it run efficiently on real robots while maintaining performance gains?

\textbf{Q3:} How do design choices—such as the use of internal features and the CFN pseudo-count network—affect overall performance?

To this end, we evaluate our framework on four simulation benchmarks with 64 tasks and 5 real-world tasks, covering three types of VLA policies.

\subsection{Simulation Experiments}
\subsubsection{Setup and Baselines}
\textbf{Benchmarks.}
We evaluate our test-time scaling framework on the \textit{Simpler} \citep{simpler}, \textit{Libero} \citep{liu2023libero}, \textit{Robotwin1.0} \citep{mu2025robotwin}, and \textit{Robotwin2.0} \citep{robotwin2} benchmarks. 
\textit{Simpler} is a real-to-sim benchmark built upon the SAPIEN simulator and the ManiSkill2 benchmark. 
It provides simulated task environments for both the WindowX and Google Robot platforms; in this work, we primarily utilize the tasks designed for WindowX. 
\textit{Libero} is a benchmark for lifelong learning in decision making (LLDM), consisting of four task suites. 
We focus on the most challenging suite, as previous works have already achieved near-perfect success rates on the others. 
\textit{Robotwin1.0} and \textit{Robotwin2.0} mainly focus on dual-arm gripper manipulation, featuring a large variety of assets and task types, and providing scripted policies for automatic data collection.

\textbf{Baselines.}
We primarily evaluate our framework on flow-matching-based VLA policies, including $\pi_{0}$ \citep{black2024pi0} and $\pi_{0.5}$ \citep{black2025pi05}. 
To demonstrate the generality of our framework, we also apply test-time scaling to the autoregressive-based \textit{OpenVLA} \citep{kim2024openvla} framework and compare it with \textit{RoboMonkey} \citep{RoboMonkey}, which is likewise a test-time scaling framework. 
\textit{RoboMonkey} builds upon \textit{LLaVA-7B} \citep{llama2} and employs preference learning to train an action verifier.
For a more comprehensive comparison, we also report the success rates of RT-1-X \citep{RTX}, Octo \citep{team2024octo}, RoboVLM \citep{robovlm}, SpatialVLA \citep{qu2025spatialvla}, and RDT \citep{liu2025rdt} on selected benchmarks.

\subsubsection{Results}

To answer \textbf{Q1:} \textit{Can our framework improve performance across diverse environments, various VLA policies, and a wide range of tasks?} We evaluate our test-time scaling framework on four benchmarks, all of which achieve the highest average success rate, as shown in Table \ref{tab:robotwin1} \ref{tab:simpler} \ref{tab:robotwin2}. Specifically, our method improves performance by 9.1\%, 7.5\%, 4.7\%, respectively.
In the Libero benchmark, as shown in Table \ref{tab:libero}, since our base policy $\pi_{05}$ has been fine-tuned and achieved near-perfect success rates on three of the suites, we only evaluate our performance on the Libero-long tasks. Even though the base policy already reaches an average success rate of 94.8\%, TACO further improves it by 1.8\%. In particular, the success rate on the \textit{Moka Pots on Stove} task increases from 68\% to 86\%. Moreover, we also evaluate our method when applied to OpenVLA in Libero. Despite only introducing a lightweight MLP pseudo-counter, it still helps OpenVLA achieve a 6.0\% improvement in average success rate. For comparison, Robomonkey, trained with a 7B VLM, only achieves a 6.7\% improvement.

\begin{wrapfigure}{r}{0.5\linewidth}
    \centering
    \begin{minipage}{0.23\textwidth}
        \centering
        \includegraphics[width=\textwidth]{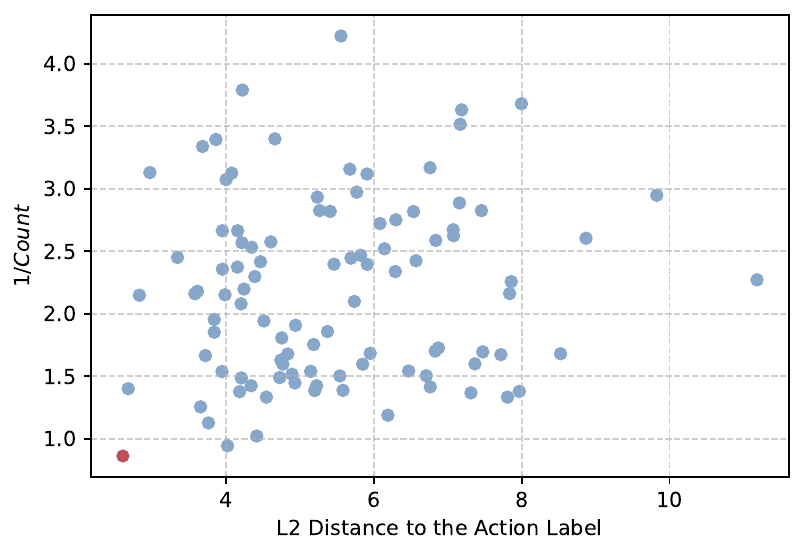}
    \end{minipage}
    \begin{minipage}{0.23\textwidth}
        \centering
        \includegraphics[width=\textwidth]{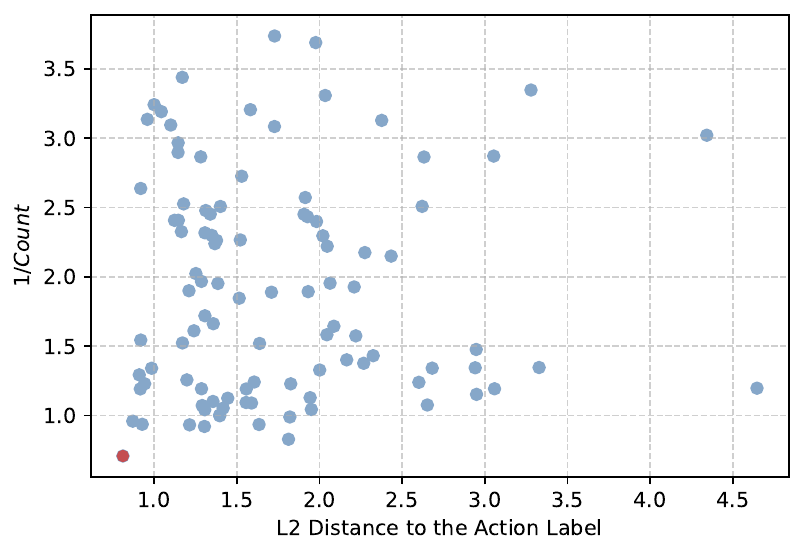}
    \end{minipage}
    \caption{We performed 100 action samples for a single observation from the test set in each of the \textit{stamp seal}(left), \textit{move can pot}(right) tasks, and scored them using the pseudo-counter. The red points indicate the actions with the highest Count values, which generally correspond to the actions with the smallest $L_2$ distance to the action labels.
    }
    \label{fig:count_dis}
\end{wrapfigure}
To further investigate the underlying reason for the effectiveness of our approach, we conducted an additional analysis. For the same test set observations, we generated 100 random noise inputs and examined the relationship between the norm of the CFN (the reciprocal of the Count) output and the $L_2$ distance between the predicted action and the ground-truth action. As shown in Figure~\ref{fig:count_dis}, the results indicate that selecting the action corresponding to the smallest CFN output norm (i.e., the largest Count) almost always yields the action closest to the ground truth, while effectively filtering out overly aggressive actions.

            

\begin{table}[t]
    \centering
    \caption{Evaluation of success rate(\%) on the RoboTwin 1.0.} 
    \label{tab:robotwin1}
    \resizebox{0.85\textwidth}{!}{
        \begin{tabular}{c|ccccc}
        \toprule
        ~ & Block Handover & Bottles Adjust & Container Place & Diverse Bottles Pick & Dual Bottles Pick Easy \\
        \midrule
        $\pi_0$ & 41.0 & 31.0 & 25.0 & 21.0 & 60.0 \\
        \rowcolor{gray!10}
        \cellcolor{white}{$\pi_0$ \textbf{+ TACO (Ours)}} & \emphTab{\bf{62.0}}{\up{$\uparrow$ 21}} & \emphTab{\bf{40.0}}{\up{$\uparrow$ 9}} & \emphTab{\bf{40.0}}{\up{$\uparrow$ 15}} & \emphTab{\bf{27.0}}{\up{$\uparrow$ 6}} & \emphTab{\bf{70.0}}{\up{$\uparrow$ 10}} \\
        \midrule
        ~ & Dual Bottles Pick Hard & Pick Apple Messy & Shoe Place & Mug Hanging Easy & Average \\  \midrule
        $\pi_0$ & 48.0 & 15.0 & 42.0 & 7.0 & 32.2 \\
        \rowcolor{gray!10}
        \cellcolor{white}{$\pi_0$ \textbf{+ TACO (Ours)}} & \emphTab{\bf{52.0}}{\up{$\uparrow$ 4}} & \emphTab{\bf{19.0}}{\up{$\uparrow$ 4}} & \emphTab{\bf{50.0}}{\up{$\uparrow$ 8}} & \emphTab{\bf{12.0}}{\up{$\uparrow$ 5}} & \emphTab{\bf{41.3}}{\up{$\uparrow$ 9.1}} \\
        \bottomrule
        \end{tabular}
    }
\end{table}

\begin{table}[t]
    \centering
    \caption{Success rate (\%) evaluated on Simpler-WindowX. Baseline results are taken from \textit{Simpler}, \textit{RoboVLM}, and \textit{SpatialVLA}. Our method achieves an average improvement of 7.5\% over $\pi_0$.}
    \label{tab:simpler}
    \resizebox{0.7\textwidth}{!}{
        \begin{tabular}{ccccc|c >{\columncolor{gray!10}}c}
        \toprule
        ~ & RT-1-X & Octo & RoboVLM & SpatialVLA & $\pi_0$ & \cellcolor{white}{$\pi_0$ \textbf{+ TACO (Ours)}} \\
        \midrule
        Spoon on Towel & 0.0 & 12.5 & 29.2 & 16.7 & 36.0 & \emphTab{{\bf 52.0}} {\up{$\uparrow$ 16}} \\
        Carrot on Plate & 4.2 & 8.3 & 25.0 & 25.0 & 42.0 & \emphTab{{\bf 52.0}} {\up{$\uparrow$ 10}} \\
        Stack Cubes & 0.0 & 0.0 & 12.5 & 29.2 & \textbf{34.0} & \emphTab{{ 30.0}} {\down{$\downarrow$ 4}} \\
        Eggplant in Basket & 0.0 & 43.1 & 58.3 & \textbf{100.0} & 80.0 & \emphTab{{\bf 88.0}} {\up{$\uparrow$ 8}}\\
        \midrule
        Average & 1.1 & 16.0 & 31.3 & 42.7 & 48.0 & \emphTab{{\bf 55.5}}{\up{$\uparrow$ 7.5}} \\
            
        \bottomrule
        \end{tabular}
    }
\end{table}

\begin{table*}[tbp]
    \centering
    \caption{Evaluation on the Robotwin2.0 benchmark. Each task is tested across 100 randomly generated scenes using 100 different seeds. For test-time scaling, the number of candidate actions is set to 50.}
    \label{tab:robotwin2}
    \resizebox{\textwidth}{!}{
        \begin{tabular}{cccccccc}
        \toprule
        ~ & Adjust Bottle & Beat Block Hammer & Blocks Ranking Size & Dump Bin Bigbin & Grab Roller & Handover Block & Handover Mic \\
        \midrule
        RDT & 81.0 & 77.0 & 0.0 & 64.0 & 74.0 & \textbf{45.0} & \textbf{90.0} \\
        $\pi_{0.5}$ & 89.0 & 69.0 & 36.0 & 86.0 & \textbf{100.0} & 24.0 & 58.0 \\
        \rowcolor{gray!10}
        $\pi_{0.5}$ \textbf{+ TACO (Ours)} & \textbf{93.0} ({\up{$\uparrow$ 4}})& \textbf{79.0} ({\up{$\uparrow$ 10}})& \textbf{40.0} ({\up{$\uparrow$ 4}})& \textbf{87.0} ({\up{$\uparrow$ 1}})& 98.0 ({\down{$\downarrow$ 2}})& 36.0 ({\up{$\uparrow$ 12}})& 63.0 ({\up{$\uparrow$ 5}})\\

        \toprule
        ~ & Lift Pot & Move Can Pot & Move Pillbottle Pad & Move Playingcard Away & Move Stapler Pad & Open Laptop & Open Microwave \\
        \midrule
        RDT & \textbf{72.0} & 25.0 & 8.0 & 43.0 & 2.0 & 59.0 & \textbf{37.0} \\ 
        $\pi_{0.5}$ & 62.0 & 42.0 & \textbf{55.0} & 85.0 & 13.0 & \textbf{67.0} & 20.0 \\
        \rowcolor{gray!10}
        $\pi_{0.5}$ \textbf{+ TACO (Ours)} & 66.0 ({\up{$\uparrow$ 4}}) & \textbf{57.0} ({\up{$\uparrow$ 15}})& 54.0 ({\down{$\downarrow$ 1}})& \textbf{88.0} ({\up{$\uparrow$ 3}})& \textbf{18.0} ({\up{$\uparrow$ 5}})& \textbf{67.0} ({\up{0}})& 21.0 ({\up{$\uparrow$ 1}})\\

        \toprule
        ~ & Pick Diverse Bottles & Pick Dual Bottles & Place A2B Left & Place A2B Right & Place Bread Basket & Place Bread Skillet & Place Burger Fries \\
        \midrule
        RDT & 2.0 & 42.0 & 3.0 & 1.0 & 10.0 & 5.0 & 50.0 \\
        $\pi_{0.5}$ & 52.0 & 72.0 & 51.0 & 39.0 & 62.0 & \textbf{62.0} & 89.0 \\
        \rowcolor{gray!10}
        $\pi_{0.5}$ \textbf{+ TACO (Ours)} & \textbf{59.0}({\up{$\uparrow$ 7}}) & \textbf{73.0}({\up{$\uparrow$ 1}}) & \textbf{56.0}({\up{$\uparrow$ 5}}) & \textbf{42.0}({\up{$\uparrow$ 3}})  & \textbf{65.0}({\up{$\uparrow$ 3}})  & 61.0 ({\down{$\downarrow$ 4}}) & \textbf{92.0}({\up{$\uparrow$ 3}})  \\
        
        \toprule
        ~ & Place Cans Plasticbox & Place Container Plate & Place Dual Shoes & Place Fan & Place Mouse Pad & Place Object Basket & Place Object Scale \\
        \midrule
        RDT & 6.0 & 78.0 & 4.0 & 12.0 & 1.0 & 33.0 & 1.0 \\
        $\pi_{0.5}$ & 68.0 & 85.0 & 23.0 & \textbf{34.0} & 16.0 & 69.0 & 45.0 \\
        \rowcolor{gray!10}
        $\pi_{0.5}$ \textbf{+ TACO (Ours)} & \textbf{72.0} ({\up{$\uparrow$ 4}}) & \textbf{87.0} ({\up{$\uparrow$ 2}}) & \textbf{28.0} ({\up{$\uparrow$ 5}}) & 32.0 ({\down{$\downarrow$ 2}}) & \textbf{24.0} ({\up{$\uparrow$ 8}}) & \textbf{78.0} ({\up{$\uparrow$ 9}}) & \textbf{52.0}({\up{$\uparrow$ 7}})  \\

        \toprule
        ~ & Place Object Stand & Place Phone Stand & Place Shoe & Put Object Cabinet & Rotate QRcode & Shake Bottle Horizontally & Shake Bottle \\
        \midrule
        RDT & 15.0 & 15.0 & 35.0 & 33.0 & 50.0 & 84.0 & 74.0 \\
        $\pi_{0.5}$ & 68.0 & 76.0 & 53.0 & 54.0 & 62.0 & \textbf{100.0} & 98.0 \\
        \rowcolor{gray!10}
        $\pi_{0.5}$ \textbf{+ TACO (Ours)} & \textbf{78.0} ({\up{$\uparrow$ 10}}) & \textbf{86.0} ({\up{$\uparrow$ 10}}) & \textbf{65.0} ({\up{$\uparrow$ 12}}) & \textbf{56.0} ({\up{$\uparrow$ 2}}) & \textbf{67.0} ({\up{$\uparrow$ 5}}) & \textbf{100.0} ({\up{0}}) & \textbf{99.0} ({\up{$\uparrow$ 1}}) \\

        \toprule
        ~ & Stack Blocks Three & Stack Blocks Two & Stack Bowls Three & Stack Bowls Two & Stamp Seal & Turn Switch & Average \\
        \midrule
        RDT & 2.0 & 21.0 & 51.0 & 76.0 & 1.0 & 35.0 & 34.6 \\
        $\pi_{0.5}$ & 43.0 & 81.0 & 64.0 & \textbf{93.0} & 26.0 & 42.0 & 59.3\\
        \rowcolor{gray!10}
        $\pi_{0.5}$ \textbf{+ TACO (Ours)} & \textbf{45.0} ({\up{$\uparrow$ 2}}) & \textbf{91.0} ({\up{$\uparrow$ 10}}) & \textbf{68.0} ({\up{$\uparrow$ 4}}) & \textbf{93.0} ({\up{0}}) & \textbf{38.0} ({\up{$\uparrow$ 12}}) & \textbf{49.0} ({\up{$\uparrow$ 7}}) & \textbf{64.0} ({\up{$\uparrow$ 4.7}}) \\
            
        \bottomrule
        \end{tabular}
    }
\end{table*}

\begin{table}[t]
    \centering
    \caption{Evaluation on the Libero-long benchmark. Our method, TACO, is applied to both Pi0.5 and OpenVLA. For the autoregressive VLA architecture, we set $temperature=1$ for action sampling. Results marked with $\mathbf{*}$ are directly reported from Robomonkey \citep{RoboMonkey}. \textit{OpenVLA (reproduced)} denotes our own reproduction results, which serves as the baseline for our TACO implementation. We observe that TACO can be effectively applied to autoregressive VLA and consistently improves performance.
    }
    \label{tab:libero}
    \resizebox{0.78\textwidth}{!}{
        \begin{tabular}{cc >{\columncolor{gray!10}}c|ccc >{\columncolor{gray!10}}c}
        \toprule
        \multirow{2}{*}{} 
        & {$\pi_{0.5}$} 
        & \cellcolor{white}{\textbf{+ TACO}} 
        & {OpenVLA*} 
        & {\textbf{+} Robomonkey*} 
        & {OpenVLA} 
        & \cellcolor{white}{\textbf{+ TACO}} \\
        & \citep{black2025pi05} & \cellcolor{white}(Ours) & \citep{kim2024openvla} & \citep{RoboMonkey} & (reproduced) & \cellcolor{white}(Ours) \\
        \midrule
        Soup and Sauce in Basket & 98.0 & \emphTab{\textbf{100.0}} {\up{$\uparrow$ 2}} & 36.0 & 59.0 & 60.0 & \emphTab{\textbf{66.0}} {\up{$\uparrow$ 6}}\\
        Cheese and Butter in Basket & 100.0 & \emphTab{96.0} {\down{$\downarrow$ 4}} & 70.0 & 79.0 & 76.0 & \emphTab{\textbf{82.0}} {\up{$\uparrow$ 6}} \\
        Turn on Stove and Place Moka & 98.0 & \emphTab{98.0} {\up{0}} & \textbf{58.0} & \textbf{58.0} & \textbf{58.0} & \emphTab{52.0} {\down{$\downarrow$ 6}} \\
        Black Bowl in Drawer & 98.0 & \emphTab{\textbf{100.0}} {\up{$\uparrow$ 2}} & 36.0 & 37.0 & 36.0 & \emphTab{\textbf{50.0}} {\up{$\uparrow$ 14}} \\
        Mugs on Plates & 98.0 & \emphTab{98.0} {\up{0}} & 42.0 & 55.0 & 32.0 & \emphTab{\textbf{50.0}} {\up{$\uparrow$ 18}} \\
        Book in Caddy & 100.0 & \emphTab{100.0} {\up{0}} & 84.0 & 86.0 & 82.0 & \emphTab{\textbf{90.0}} {\up{$\uparrow$ 8}} \\
        Mug and Pudding on Plate & \textbf{96.0} & \emphTab{92.0} {\down{$\downarrow$ 4}} & 48.0 & 59.0 & \textbf{60.0} & \emphTab{54.0} {\down{$\downarrow$ 6}} \\
        Soup and Cheese in Basket & 94.0 & \emphTab{\textbf{100.0}} {\up{$\uparrow$ 6}} & 56.0 & 62.0 & 70.0 & \emphTab{\textbf{80.0}} {\up{$\uparrow$ 10}} \\
        Moka Pots on Stove & 68.0 & \emphTab{\textbf{86.0}} {\up{$\uparrow$ 16}} & 26.0 & 26.0 & 20.0 & \emphTab{\textbf{28.0}} {\up{$\uparrow$ 8}} \\
        Mug in Microwave & \textbf{98.0} & \emphTab{96.0} {\down{$\downarrow$ 2}} & 42.0 & 44.0 & 46.0 & \emphTab{\textbf{48.0}} {\up{$\uparrow$ 2}} \\
        \midrule
        Average & 94.8 & \emphTab{\textbf{96.6}}{\up{$\uparrow$ 1.8}} & 49.8 & 56.5 & 54.0 & \emphTab{\textbf{60.0}} {\up{$\uparrow$ 6}}\\
        \bottomrule
        \end{tabular}
    }
\end{table}

\subsection{Real-World Experiments}
\subsubsection{Setup}

We build our real-world experimental setup using a \textsc{RealMan75} dual-arm robot in an office-like environment. The setup includes five tasks involving various common objects such as books, pens, a cabinet, a charger, a laptop, and a paper ball. The five tasks are as follows: \textit{Receive Book}, where the robot receives a book handed over by a human and places it back on the bookshelf; \textit{Storage Charger}, where it picks up the charger head, places it inside the cabinet, and closes the cabinet door; \textit{Paper and Pen}, where it picks up the paper ball and the pen and places them into the trash bin and pen holder, respectively; \textit{Laptop}, where it closes the laptop lid, unplugs the charger from the socket, and stores it back in the cabinet; and \textit{Pick Books}, where the left and right arms each pick up and lift a book simultaneously.

These tasks cover a wide range of skills and scenarios, including human-robot interaction, dual-arm coordination, and long-horizon task execution, providing a comprehensive evaluation of the policy's general capabilities. For data collection, we perform teleoperation to record 100 episodes for each task. The collected data are then used to fine-tune the base policy $\pi_{0}$, upon which we deploy our proposed test-time scaling framework.
In the real-world experiments, we simultaneously sample 30 action chunks at each decision step, with each action chunk having a length of 20.

\subsubsection{Results}
We address \textbf{Q2:} \textit{What is the additional time cost introduced by our test-time scaling method? Can it run efficiently on real robots while maintaining performance gains?} 

\begin{figure}[t]
    \centering
    \begin{minipage}{0.56\linewidth}
        \centering
        \includegraphics[width=\linewidth]{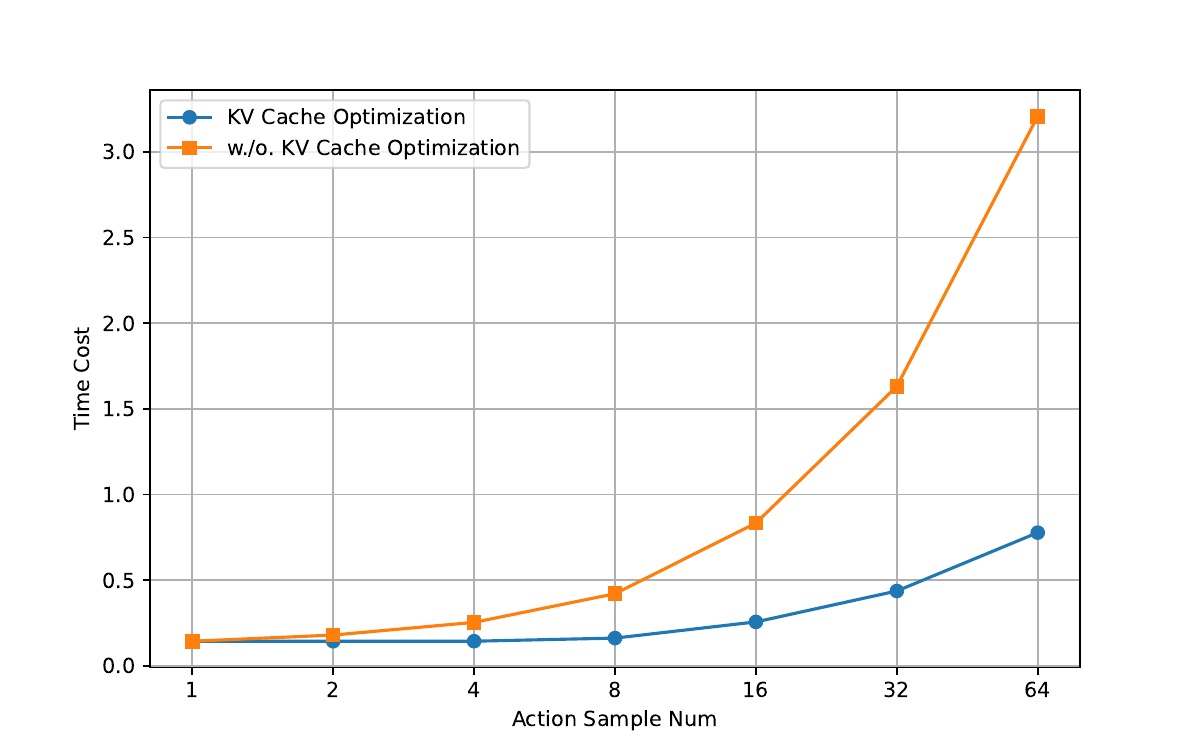}
        \label{fig:time_cost}
    \end{minipage}%
    \hfill
    \begin{minipage}{0.43\linewidth}
        \centering
        \resizebox{0.9\linewidth}{!}{
            \includegraphics[width=\linewidth]{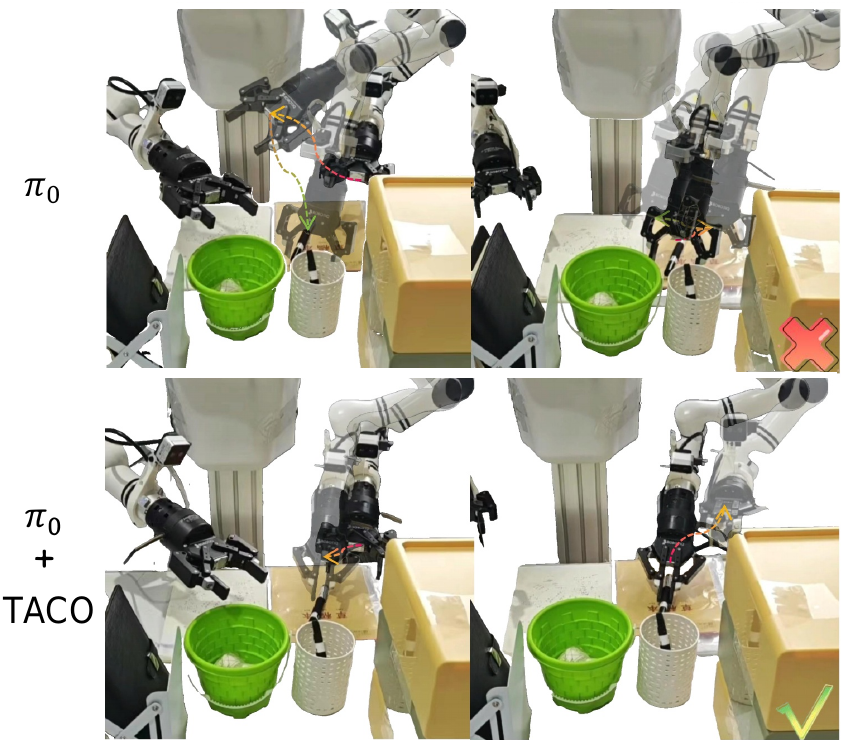}
        }
        \label{fig:real_compar}
    \end{minipage}%

    \caption{
        (Left) Efficiency improvement from KV cache optimization. For each number of action samples, we repeated the inference 50 times and reported the average inference time. 
        (Right) A key moment from the \textit{Paper and Pen} task, where the robot needs to pick up a marker. The base policy $\pi_0$ often falls into a suboptimal mode, causing the arm to swing back and forth and hesitate before grasping, which leads to task failure. In contrast, our method enables $\pi_0$ to completely avoid such oscillations and hesitation during grasping.
        }
    \label{fig:real_exp}
\end{figure}

To intuitively demonstrate the time cost of parallel inference over multiple actions, we first compare the inference latency of two settings: (1) direct parallel inference of multiple actions, and (2) our optimized implementation that shares the observation $k, v$ cache across actions. As shown in Figure \ref{fig:real_exp} (left), we evaluate the policy $\pi_{0}$ on a single RTX 4090 GPU. The results show that the $k, v$ caching optimization significantly accelerates inference, especially when the number of parallel actions is large. When sampling 32 actions simultaneously, the inference time is reduced by 73.2\% compared to naïve multi-batch parallel inference. Therefore, our method remains efficient and low-latency when deployed on real robots.



\begin{wrapfigure}{r}{0.45\linewidth}
    \centering
    \resizebox{0.44\textwidth}{!}{%
        \includegraphics{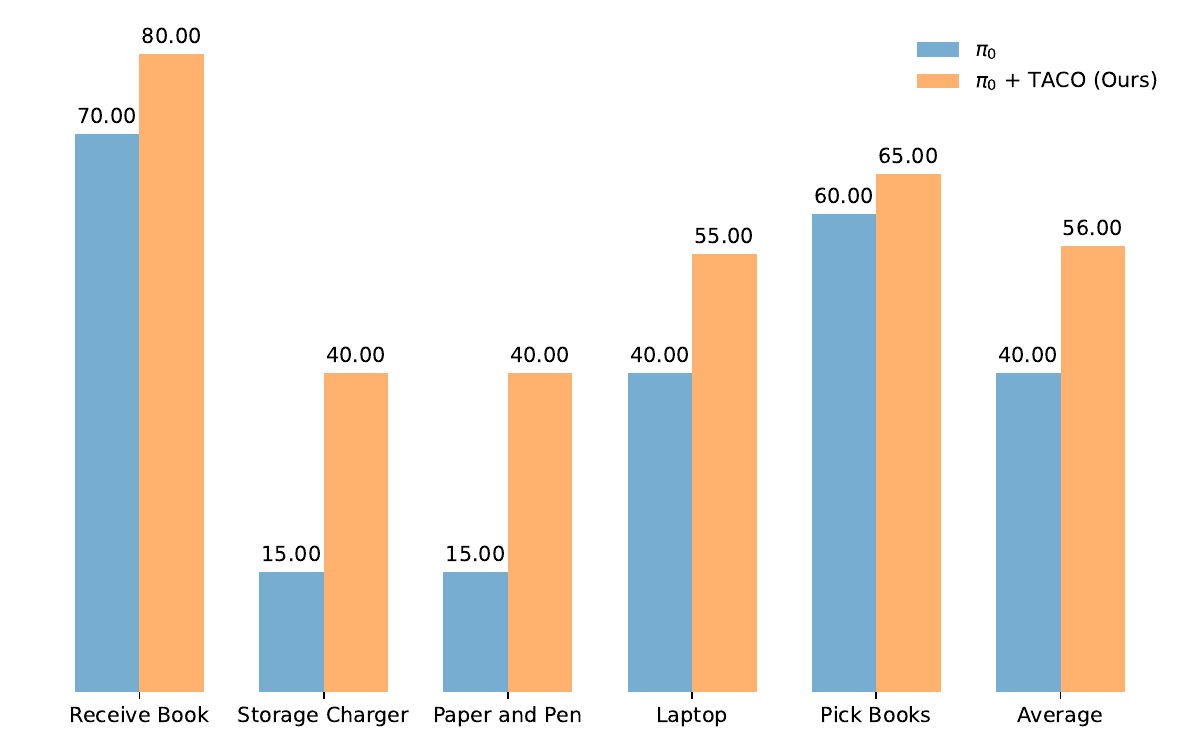}
    }
    \caption{Success rate (\%) in real-world experiments. On our self-built \textit{RealMan 75} dual-arm robotic platform, applying the proposed test-time scaling method improves both task accuracy and safety, resulting in an average 16\% increase in success rate.}
    \label{fig:real_exp_sucrate}
\end{wrapfigure}

The success rates in real-world experiments are shown in Figure~\ref{fig:real_exp_sucrate}. Our method achieves an average improvement of 16\% over the base policy, with particularly notable gains in long-horizon tasks. 
Specifically, we observe a 25\% improvement in the \textit{Paper and Pen} task and a 15\% improvement in the \textit{Laptop} task. Based on our observations, the base policy typically fails in two ways: (1) imprecise grasping positions, and (2) suboptimal teleoperation data quality. 

Our scaling framework improves action precision, consistent with the analysis observed in simulation experiments.
More importantly, our approach effectively mitigates policy failures caused by imperfect teleoperation data. As shown in Figure \ref{fig:real_exp} (right), during the \textit{Pen Grasping} stage, the operator's varying grasp poses and timings lead to large state differences between grasping modes, resulting in a sparse observation–action distribution in the expert dataset. Consequently, the base policy often exhibits unstable behaviors such as failing to close the gripper properly, re-opening after grasping, or oscillating between two grasping modes. With our test-time scaling method, these issues are largely eliminated: the gripper consistently closes correctly, and the robot avoids suboptimal modes. We attribute this improvement to the filtering effect of our scaling framework, which suppresses suboptimal trajectories introduced by noise in the teleoperated dataset.



\begin{table}[tbp]
    \centering
    \caption{We conducted an ablation study on the CFN network, the method for obtaining features, and the feature processing approach.}
    \label{tab:ablation}
    \resizebox{0.68\textwidth}{!}{
        \begin{tabular}{ccccc}
        \toprule
        ~ & Block Handover &  Container Place & Diverse Bottles Pick & Average \\
        \midrule
        $\pi_0$ \textbf{+ TACO (Ours)} & \textbf{62.0} & \textbf{40.0} & 27.0 & \textbf{43.0} \\
        w./o. CFN & 52.0 & 32.0 & 26.0 & 36.7 \\
        w./o. Feature Scaling & 51.0 & 37.0 & \textbf{30.0} & 39.3 \\
        w./o. Internal Feature & 48.0 & 33.0 & 23.0 & 34.7 \\
            
        \bottomrule
        \end{tabular}
    }
\end{table}

\subsection{Ablation Studies and Analyses}
To answer \textbf{Q3:} \textit{How do design choices—such as the use of internal features and the CFN pseudo-count network—affect overall performance?} We designed a set of ablation experiments based on $\pi_0$ on the RobotWin benchmark. As shown in Table \ref{tab:ablation}, the experiments include: (1) replacing CFN pseudo-counting of good features with a direct fit of the Euclidean distance between the predicted action from the features and the ground-truth action; (2) applying CFN for pseudo-counting without performing feature scaling on Internal Features; (3) using a CNN-based image encoder and an MLP-based action encoder to obtain features instead of relying on Internal Features.

Directly fitting the mapping from features to action errors increases the learning difficulty, as the model must capture not only optimal features but also suboptimal and poor ones. During test-time scaling, only the selection of the optimal action matters. Our experimental analysis suggests that the observed performance drop is caused by overfitting on the training set, leading to reduced generalization. When separate encoders are used instead of Internal Features and trained jointly with CFN, the resulting features tend to be highly similar, making accurate pseudo-counting difficult and causing a significant performance drop. This approach also reduces efficiency, as it fails to leverage VLA’s inherent understanding of scenes and actions.

\section{Conclusion}

In this work, we introduce \textbf{TACO}, a test-time-scaling (TTS) framework that employs a lightweight pseudo-count estimator as a high-fidelity verifier for action chunks. 
TACO is compatible with various VLA models and consistently delivers substantial improvements over the base policy across a wide range of real-world and simulated experiments. 
Despite certain limitations—such as its inability to evaluate newly synthesized actions produced purely by perturbation and its reliance on the representational capability of the underlying VLA—our approach demonstrates that lightweight test-time scaling can effectively achieve anti-exploration and mitigate the instability and performance degradation caused by distribution shift in action prediction.

\clearpage

\bibliographystyle{plainnat}
\bibliography{paper}

\clearpage
\appendix

\section{Algorithmic Description of TACO during Inference}\label{app:algo}

As shown in Algorithm \ref{alg:tt_anti_exploration}, we provide an algorithmic description for the action-filtering procedure during inference time, illustrating how the CFN verifier leverages internal representations to obtain pseudo-counts, as well as the workflow of the KV Cache Optimization.
Here, we take flow-based VLAs as representative examples.

\begin{algorithm}[h]
\caption{Anti-Exploration Sampling for TTS}
\label{alg:tt_anti_exploration}
\KwIn{
    Observation $\mathbf{o}_t$; instruction $l$; \\
    VLA policy $\pi_\theta$; CFN verifier $f_\phi$; \\
    Number of candidates $M$.
}
\KwOut{Selected action chunk $\mathbf{a}^*_{t:t+H}$}

\BlankLine
Compute context KV cache $\mathcal{C}_\text{KV} = \text{ContextForward}(\mathbf{o}_t, l)$ in $\pi_\theta$\;
Sample $M$ noise candidates $\{\epsilon_i\}_{i=1}^M \sim \mathcal{N}(0,I)$\;
\For{$i = 1$ \KwTo $M$}{
    Compute the predicted clean action chunk and last hidden state $( \hat{\mathbf{a}}^{(i)}_{t:t+H},\, h_\theta^{(i)} )
        = \pi_\theta(\epsilon_i, \mathcal{C}_\text{KV}),\text{ i.e., } \pi_\theta(\mathbf{o}_t, l; \epsilon_i)$\;
}

\For{$i = 1$ \KwTo $M$}{
    Compute the pseudo-count of the generated action chunk $\hat{N}_{\mathcal{D}_\text{sft}} (\mathbf{o}_t,l,\hat{\mathbf{a}}^{(i)}_{t:t+H}) = \frac{1}{\| f_\phi(h_\theta^{(i)}) \|^2}$\;
}

Select $i^* = \arg\max_i \hat{N}_{\mathcal{D}_\text{sft}} (\mathbf{o}_t,l,\hat{\mathbf{a}}^{(i)}_{t:t+H})$\;

\Return $\mathbf{a}^*_{t:t+H} = \hat{\mathbf{a}}^{(i^*)}_{t:t+H}$

\end{algorithm}

\section{Algorithmic Description of TACO during Training}
Algorithm \ref{alg:pseudo_count} summarizes how we train the coupled pseudo-count estimator on top of a VLA.
The goal is to efficiently obtain in-distribution, high-fidelity features by repeatedly querying the VLA under different noise conditions, then selecting the representation that best aligns with the clean action. Training a lightweight CFN on top of these coupled features yields a pseudo-count estimator that reliably reflects how close a candidate action lies to the support of the dataset $\mathcal{D}_\text{sft}$, enabling effective anti-exploration during inference. Here, we also take flow-based VLAs as representative examples.

\newcommand{\CommentLine}[1]{\textcolor{gray}{//  \texttt{#1}}}

\begin{algorithm}[h]
\caption{Coupled Pseudo-Count Estimator Training for VLAs}
\label{alg:pseudo_count}
\KwIn{
    SFT dataset $\mathcal{D}_\text{sft} = \{(\mathbf{o}, l, \mathbf{a})\}$; \\
    VLA policy $\pi_\theta$; CFN head $f_\phi$; learning rate $\eta$; \\
    Noise schedule $\{\sigma_i\}_{i=1}^N$; number of queries $N$; \\
    Number of training steps $T$; batch size $B$.
}
\KwOut{Trained CFN $f_\phi$; high-fidelity feature set $\mathcal{D}_h$}

\BlankLine

\CommentLine{High-Fidelity Feature Search}

Initialize feature set: $\mathcal{D}_h \leftarrow \emptyset$\;

\ForEach{$(\mathbf{o}, l, \mathbf{a}) \in \mathcal{D}_\text{sft}$}{
    \For{$i = 1$ \KwTo $N$}{
        Corrupt the clean action $\mathbf{a}_{\sigma_i} = \sigma_i \mathbf{a} + (1 - \sigma_t) \epsilon$ with $\epsilon \sim \mathcal{N} (0 ,I)$\;
        Compute the predicted clean action chunk and last hidden state $(\mathbf{a}^{(i)}_{\text{pre}},\, h^{(i)}_\theta)
            = \pi_\theta(\mathbf{o}, l, \mathbf{a}_{\sigma_i})$\;
        Calculate the error between the prediction and the ground truth label $e_i \leftarrow \| \mathbf{a}^{(i)}_{\text{pre}} - \mathbf{a} \|_2$\;
    }
    Select $i^\star = \arg\min\nolimits_i e_i$\;
    Add the internal feature of the closest prediction into the feature set $\mathcal{D}_h \leftarrow \mathcal{D}_h \cup \{h^{(i^\star)}_\theta\}$\;
}

\BlankLine
\CommentLine{CFN Training on Coupled Features}\;

\For{$t = 1$ \KwTo $T$}{
    Sample mini-batch $\{ h_k \}_{k=1}^B \sim \mathcal{D}_h$\;
    Compute $\mathcal{L}_\text{CFN}$ according to Eq.~\eqref{equ:cfn_loss}\;
    Update CFN parameters: $\phi \leftarrow \phi - \eta \nabla_\phi \mathcal{L}_\text{CFN}$\;
}

\Return{$f_\phi,\, \mathcal{D}_h$}

\end{algorithm}

\section{Additional Preliminary}
\noindent\textbf{Flow Matching.}
Flow matching \citep{liu2023flow,lipman2023flow, lipman2024flowguide} is a generative modeling framework that learns a continuous velocity field to transform a simple prior distribution into a target data distribution. Similar to diffusion models, it defines a time-dependent transformation; however, the dynamics are governed by an ordinary differential equation (ODE) rather than a stochastic process.

Formally, let \( p(x) \) denote the target distribution on \( \mathbb{R}^k \), and let \( v(u, x): [0, 1] \times \mathbb{R}^k \to \mathbb{R}^k \) be a time-dependent velocity field. Its flow \( \psi(u, x) \) is defined by
\begin{align}
    \frac{\mathrm{d}}{\mathrm{d}u}\psi(u, x) = v(u, \psi(u, x)), \quad \psi(0, x) = x. 
\end{align}
By integrating this ODE from \( u=0 \) to \( u=1 \), samples from a simple prior (e.g., \( \mathcal{N}(0, I) \)) are continuously mapped to the target distribution. 

Flow matching provides a stable and efficient alternative to diffusion-based generation and offers a principled continuous-time formulation that can be extended to policy learning or trajectory generation in robotic systems.

\section{Training Implementation Details}

\textbf{Internal Representation Details.}
For the two policies $\pi_{0}$ and $\pi_{0.5}$, we build the training and deployment pipeline based on the \textit{Lerobot} framework \citep{cadene2024lerobot}. We first modify the model’s action-output function so that it simultaneously returns both the action and the internal representation. Concretely, we extract the first action token from the final hidden layer. Since the final hidden layer encodes an abstract representation of the entire input sequence—and, through the self-attention mechanism, the last token’s vector contains contextual information of the whole sequence—we treat this vector as a compact representation of the input.
Because the action expert used in both $\pi_{0}$ and $\pi_{0.5}$ adopts a bidirectional-attention Transformer architecture, in practice we take the first token of the final hidden layer inside the action expert as the internal representation. Its dimensionality is 1024.
For \textit{OpenVLA}, we instead take the last token of its final hidden layer as the internal representation, whose dimensionality is 4096. We set the temperature to 1 and leverage the inherent randomness in decoding to sample diverse actions and features.

\begin{figure*}[t]
    \centering
    \resizebox{0.9\textwidth}{!}{%
        \includegraphics{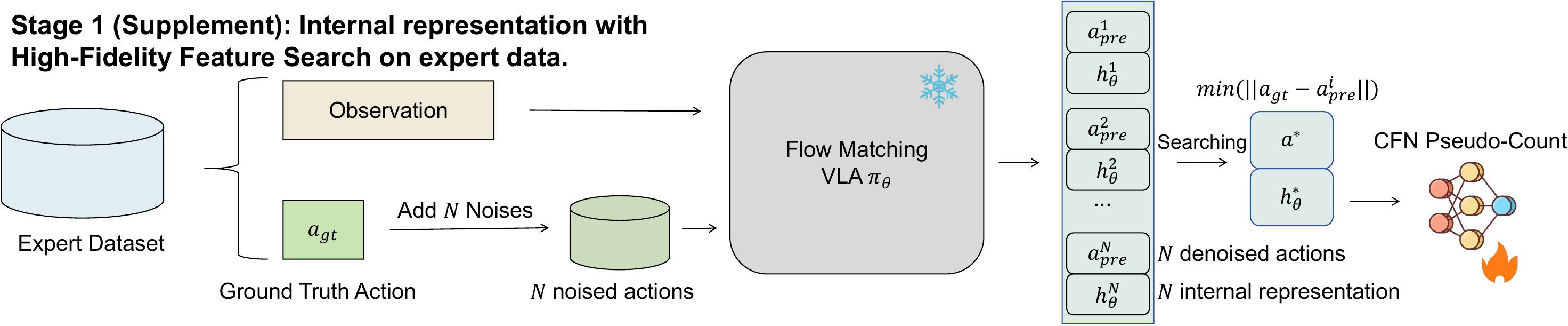}
    }
    \caption{During training, each internal representation undergoes a search process that selects the best one from $N$ candidates to maximally preserve the information of the ground-truth action.}
    \label{fig:stage1_searching}
\end{figure*}

\textbf{Base Policy and Representation Dataset.}
Taking \textit{Robotwin1.0} \citep{mu2025robotwin} as an example, we first collect 50 trajectories for each task using the provided scripted policies to construct the dataset. 
Expert data collected via scripted policies can typically avoid a large amount of human-induced noise, enabling the construction of high-quality datasets at low cost \citep{humanoidgen}.
We then adopt the $\pi_0$ base model released by LeRobot on Hugging Face, and perform full-parameter fine-tuning on all task data using an NVIDIA H100 with a batch size of 48 for 30k steps, resulting in the base policy.
We inject 10\% noise into the ground-truth actions and apply a single denoising step using the $\pi_0$ action expert to obtain the internal representations. 
As shown in Figure \ref{fig:stage1_searching}, 
for each triplet $(\mathbf{o}, l, \mathbf{a})$, we additionally sample $N = 50$ noise instances to generate:
\(
\{ (\mathbf{a}^{(i)}_{\text{pre}}, h^{(i)}_\theta) \}_{i=1}^N,
\)
and select the sample $\mathbf{a}^{(*)}_{\text{pre}}$ that is closest to the ground-truth action, which is then stored as input for subsequent training.

For \textit{Robotwin2.0} \citep{robotwin2}, we similarly use the LeRobot $\pi_0$ base model from Hugging Face and fine-tune it on an NVIDIA H100 with a batch size of 32 for 10k steps across all tasks. When obtaining internal representations, 
we empirically set the noise level to 100\% relative to the ground-truth actions. All other settings remain unchanged.

For \textit{Libero} \citep{liu2023libero}, we directly use the official LeRobot model \texttt{pi0\_libero\_finetuned} as well as the open-source \texttt{openvla-7b-finetuned-libero-10} available on Hugging Face as our base policies.

For \textit{Simpler} \citep{simpler}, we use the \textit{Bridge V2} dataset and perform full-parameter fine-tuning of $\pi_0$ on 8 NVIDIA H100 GPUs with a batch size of 256, which we take as the base policy.

\textbf{Training Parameters}
We take the training parameters on RobotWin2.0 as an example, and the remaining training configurations are largely similar. 
The training process of the CFN follows the settings listed in Table \ref{tab:cfn_training_params}. 
The Adam optimizer is selected for parameter updates due to its efficiency in handling sparse gradients;
The OneCycleLR scheduler is employed to dynamically adjust the learning rate: it linearly increases the learning rate from the initial value ($10^{-4}$) to the maximum value ($10^{-3}$) in the first half of training, and then cosine-anneals the learning rate to a lower value in the second half, which helps to balance convergence speed and generalization performance;
Gradient accumulation is adopted with a step size of 2, i.e., gradients from 2 consecutive batches are accumulated before updating the model parameters. This effectively simulates a larger batch size while avoiding excessive GPU memory consumption.

\begin{table}[htbp]
  \centering
  \caption{Training parameters of CFN}
  \label{tab:cfn_training_params}
    \resizebox{0.48\textwidth}{!}{
        \begin{tabular}{lc}
        \toprule
        Parameter Name & Value \\
        \midrule
        Batch Size & 512 \\
        Action Chunk Size & 50 \\
        CFN Output Dimension (\texttt{cfn\_output\_dim}) & 20 \\
        CFN Hidden Dimension (\texttt{cfn\_hidden\_dim}) & 1536 \\
        Total Training Epochs (\texttt{total\_epochs}) & 16 \\
        Optimizer & Adam \\
        Initial Learning Rate & $10^{-4}$ \\
        Learning Rate Scheduler & OneCycleLR \\
        Maximum Learning Rate of Scheduler & $10^{-3}$ \\
        Annealing Strategy & Cosine Annealing \\
        Gradient Accumulation Steps (\texttt{grad\_accum\_steps}) & 2 \\
        \bottomrule
      \end{tabular}
    }
\end{table}


\section{Training Procedure Details}

\subsection{Network Architecture of CFN}
The Coin Flipping Network (CFN) adopted in this work is a lightweight multi-layer perceptron (MLP), designed for efficient feature processing and target prediction. 
Its overall structure consists of four core components: a feature scaling module, an input projection layer, a stack of one or more MLP blocks, and an output projection layer. Detailed implementations are as follows:

\textbf{Feature Scaling}
Before entering the network, all input features are scaled by a constant factor (set to 10 in this work) to enlarge the inter-feature distances. 
This operation helps reduce the interference between features during subsequent processing and improves the stability of feature discrimination.

\textbf{MLP Block}
Each MLP block serves as the basic computational unit, employing a residual connection mechanism to mitigate the vanishing gradient problem during training. The structure of a single MLP block includes:
\begin{itemize}
    \item Two fully connected (FC) layers, where the first layer maps the input feature dimension to 4 times the hidden dimension, and the second layer projects it back to the original hidden dimension;
    \item The GELU (Gaussian Error Linear Units) activation function is used to introduce non-linearity;
    \item A Dropout layer for regularization to avoid overfitting;
    \item A LayerNorm layer applied after the residual connection to stabilize the training process.
\end{itemize}
Weight initialization follows the Xavier uniform distribution, and biases are initialized to zero.

The specific workflow of the CFN network is as follows: 
\textit{Feature Scaling}: All input features are multiplied by 10 to broaden their distribution range and improve feature separability; 
\textit{Input Projection}: The scaled features are projected into the predefined hidden dimension using a fully connected (FC) layer, followed by GELU activation and LayerNorm normalization; 
\textit{Feature Enhancement}: The normalized features are then passed through an MLP block for deep feature extraction and enhancement; 
\textit{Output Projection}: Finally, the enhanced features are mapped to the target output dimension via the last FC layer.


\subsection{CFN Loss.}
In our setting, the CFN is trained not on raw states but on the high-fidelity feature set
$\mathcal{D}_h = \{h^{(i^\star)}_\theta\}$ extracted from the frozen VLA backbone.
Following the same formulation as Eq.~\eqref{equ:cfn_loss}, the learning objective becomes
\begin{align}\label{equ:cfn_feat_loss}
    \mathcal{L}_{\text{CFN}}(\phi)
    = \mathbb{E}_{(h_\theta, \mathbf{c}) \sim \mathcal{D}_h}
    \big[\, \| f_\phi(h_\theta) - \mathbf{c} \|^2 \,\big],
\end{align}
where each feature $h_\theta$ is paired with a randomly sampled Rademacher vector
$\mathbf{c} \sim \{-1,1\}^d$.
Minimizing this distance-based loss encourages $f_\phi$ to approximate the expected coin-flip outcome
associated with each feature embedding.
As in the original CFN formulation, the norm of the optimal solution,
$\| f_\phi(h_\theta) \|^2$, implicitly encodes the inverse visitation frequency of $h_\theta$.
Hence, Eq.~\eqref{equ:pseudo_count} can be directly interpreted as a
distance-derived pseudo-count:
features that are frequently observed during training yield smaller reconstruction errors and larger norms,
whereas novel or out-of-support features correspond to larger distances in the CFN output space and smaller pseudo-count estimates.
This formulation allows the CFN to act as a smooth verifier over the feature manifold without requiring explicit density estimation.

\subsection{Random Prior Initialization}
To implement a low-count initialization for unseen features, we decompose the CFN output into a trainable component and a frozen random prior:
\begin{equation}
    f_\phi(h) = \hat{f}_\phi(h) + f_{\text{prior}}(h),
\end{equation}
where $f_{\text{prior}}$ is a randomly initialized, non-trainable network. We normalize the prior using running statistics so that each output dimension satisfies
\begin{equation}
    \mathbb{E}_{h \sim \mathcal{D}_h}\!\left[f_{\text{prior}}^{(i)}(h)^2\right] = 1,
    \qquad \forall\, i \in \{1,\dots,d\},
\end{equation}
which yields the expected squared norm
\begin{equation}
    \mathbb{E}_{h \sim \mathcal{D}_h}\!\left[\|f_{\text{prior}}(h)\|^2\right] = d.
\end{equation}

For features $h$ in regions where the learned component has not yet adapted, we have $\hat{f}_\phi(h) \approx 0$, and thus
\begin{equation}
    \| f_\phi(h) \|^2 \approx \| f_{\text{prior}}(h) \|^2 \approx d.
\end{equation}
Since our pseudo-count is defined up to a proportionality constant, we calibrate it such that
\begin{equation}
    \hat{N}_{\mathcal{D}_\text{sft}}(h) \approx \frac{d}{\|f_\phi(h)\|^2} \approx 1.
\end{equation}
In this way, unseen regions of the representation space are initialized with a pseudo-count of~1, and the influence of the random prior naturally diminishes as $\hat{f}_\phi$ is updated during training.

\section{Additional Simulation Experiment Details}

\begin{figure*}[t]
    \centering
    \resizebox{0.9\textwidth}{!}{%
        \includegraphics{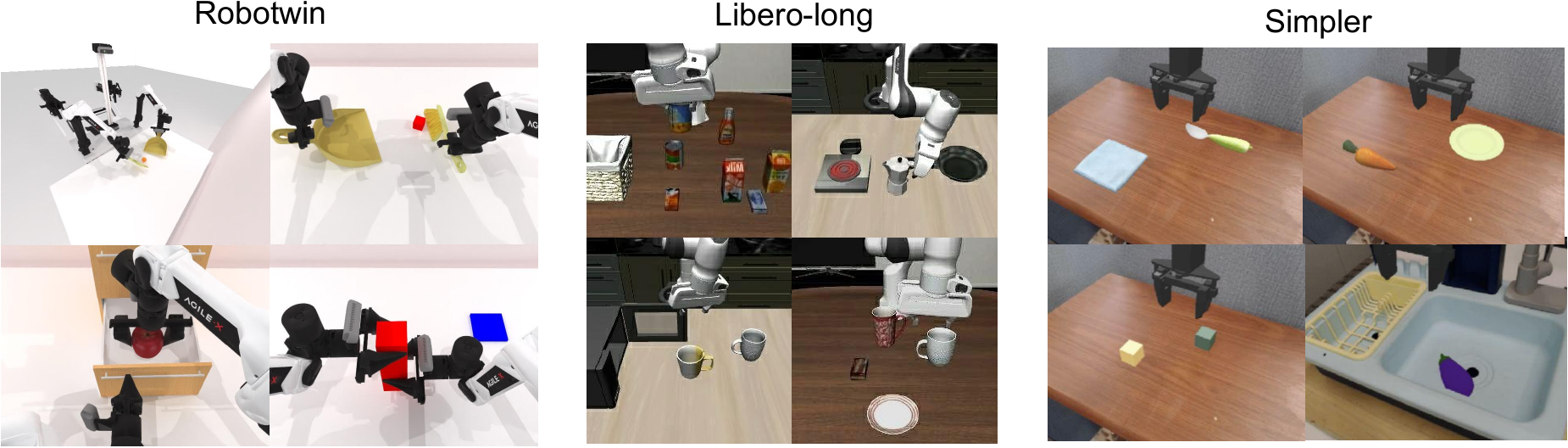}
    }
    \caption{Illustration of the benchmarks.}
    \label{fig:benchs}
\end{figure*}

\textbf{Benchmarks.}
The schematic illustration of each benchmark task is shown in Figure \ref{fig:benchs}.
\textit{RoboTwin 1.0} \citep{mu2025robotwin}: One of the earlier dual-arm robot manipulation benchmarks based on generative digital twins, providing both real-world and simulated demonstration data for dual-arm tasks.
\textit{RoboTwin 2.0} \citep{robotwin2}: A scalable extension of RoboTwin that supports automatic large-scale data generation, strong domain randomization across clutter, lighting, background, tabletop height and language, and evaluation across multiple robot embodiments for robust generalization.
\textit{LIBERO} \citep{liu2023libero}: A lifelong robot learning benchmark designed to study knowledge transfer in related objects/skills, comprising 130 manipulation tasks across four task suites and supporting research in multi-task and lifelong learning for robotics. 
\textit{SIMPLER} \citep{simpler}: A simulation-based evaluation environment set (Simulated Manipulation Policy Evaluation for Real Robot Setups) built on SAPIEN + ManiSkill2, aimed at providing scalable, reproducible simulation evaluations of manipulation policies that correlate well with real-world performance.

\begin{figure*}[t]
    \centering
    \resizebox{\textwidth}{!}{%
        \begin{minipage}{\textwidth}
            \centering
            \begin{subfigure}[t]{0.32\textwidth}
                \includegraphics[width=\textwidth]{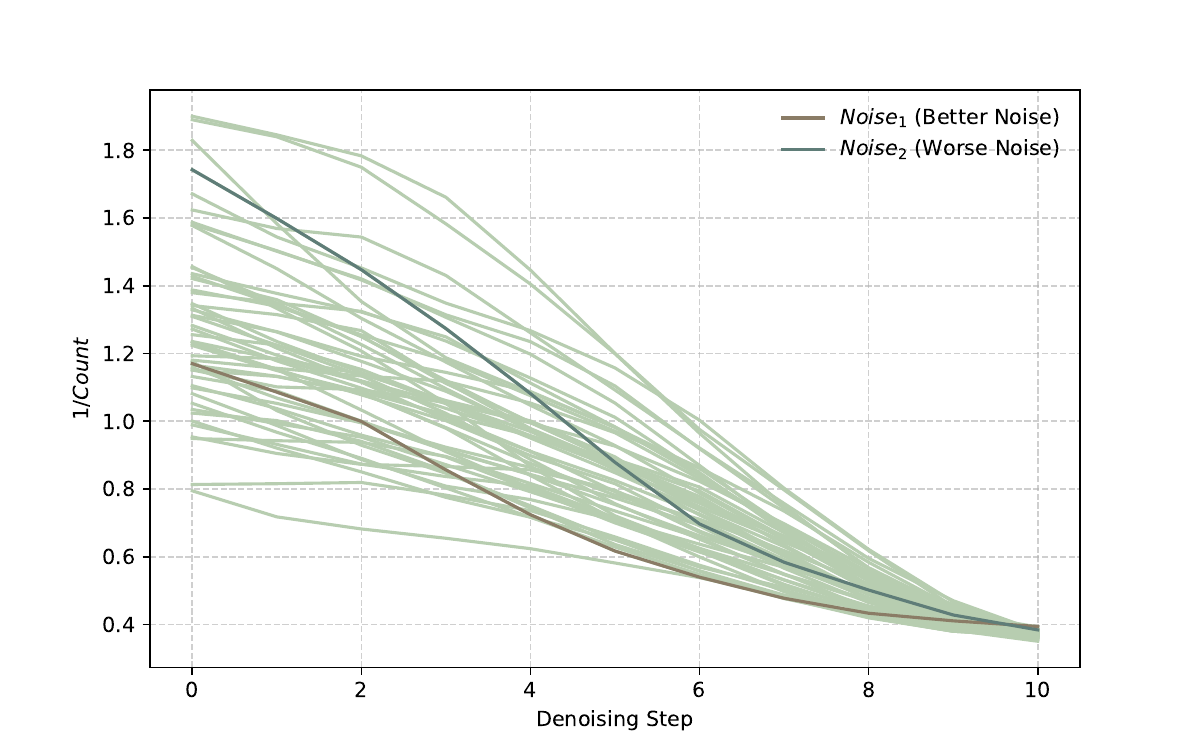}
                \caption{CNN + MLP Encoder}
            \end{subfigure}
            \hfill
            \begin{subfigure}[t]{0.32\textwidth}
                \includegraphics[width=\textwidth]{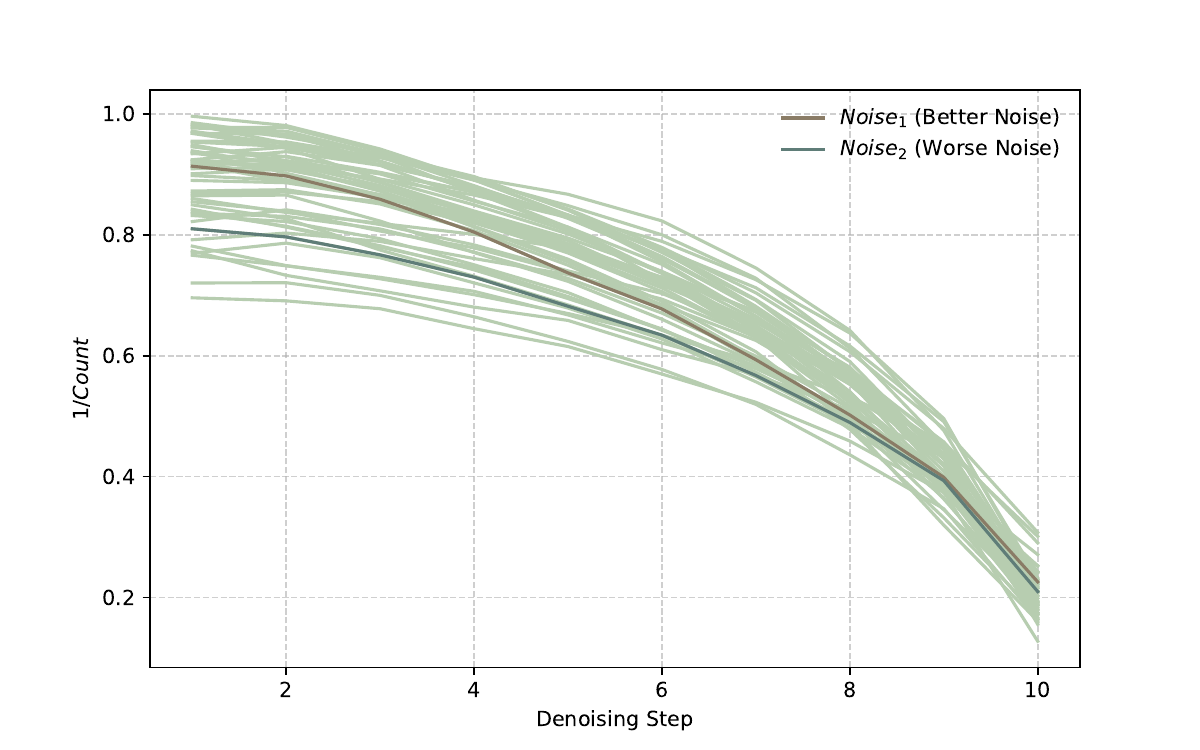}
                \caption{Internal Representation}
            \end{subfigure}
            \hfill
            \begin{subfigure}[t]{0.32\textwidth}
                \includegraphics[width=\textwidth]{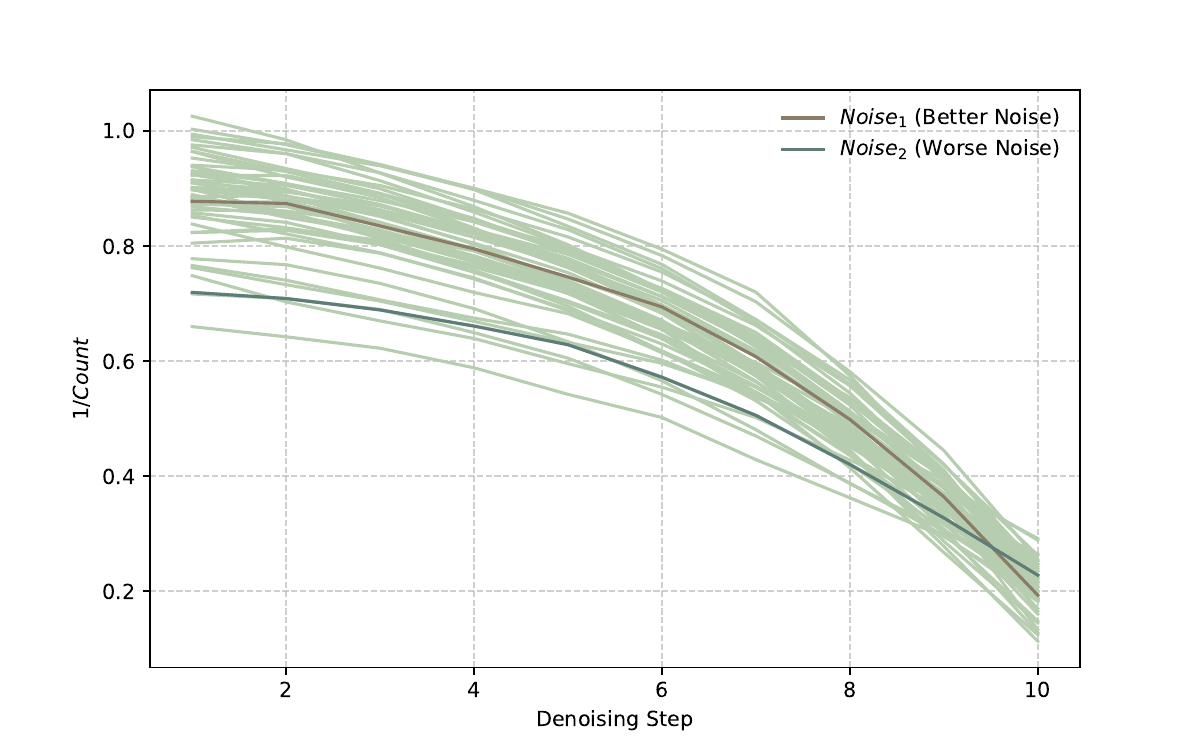}
                \caption{Internal Representation  with High-Fidelity Feature Search}
            \end{subfigure}
        \end{minipage}%
    }

    \caption{Comparison of the Count evolution during the action denoising process under different representation extraction methods, illustrated with the Block Handover task as an example.}
    \label{fig:repre_compar}
\end{figure*}

\textbf{Further Experimental Exploration on Representations.}
We conducted extensive experiments using various methods for obtaining representations, including employing CNN and MLP as encoders (jointly optimized with the CFN parameters), utilizing internal representations, and combining internal representations with High-Fidelity Feature Search. As shown in Figure \ref{fig:repre_compar}, we visualize how $1/\text{Count}$ evolves during the denoising process across 50 different initial noise configurations in the Block Handover task under different representation settings. 

When using CNN and MLP as encoders, the pseudo-counter gradually fails to distinguish between different noise samples as the number of denoising steps increases. Eventually, the values converge to almost a single point, and the counter even reaches its maximum value before complete denoising, indicating that the counter is overly tolerant to actions. 

In contrast, when using internal representations, the counter maintains a high degree of discrimination even after the noise has been completely removed. However, it shows limited distinction between better and worse noise samples. We attribute this to the fact that noise injection during the internal representation extraction process partially corrupts the ground-truth action information. 

Finally, when internal representations are combined with High-Fidelity Feature Search, the search procedure effectively preserves the information of the ground-truth action by exploring the predicted action space. As a result, the model successfully distinguishes between better and worse noise samples once denoising is completed.

\textbf{More experiments on the choice of pseudo-counter. }
To demonstrate the role and generalization ability of CFN, 
we compare it with RND \citep{nikulin2023anti} as a replacement pseudo-counter on a subset of tasks. 
RND employs a fixed random target network and a learned predictor whose estimation error serves as an intrinsic reward, enabling a lightweight and robust measure of feature novelty. 
We use this novelty as the pseudo-count value, where higher novelty corresponds to a smaller pseudo-count. 
All other components remain unchanged: using the same internal representations, we select the action whose representation exhibits the smallest novelty as the final output. 
Table \ref{tab:counters} reports the success rate comparison. 
As shown, using CFN as the pseudo-counter yields consistently better average performance.

\begin{table}[tbp]
    \centering
    \caption{Performance comparison of final success rates (\%) using different pseudo-counters}
    \label{tab:counters}
    \resizebox{0.68\textwidth}{!}{
        \begin{tabular}{ccccc}
        \toprule
        ~ & Block Handover &  Container Place & Diverse Bottles Pick & Average \\
        \midrule
        $\pi_0$ & 41.0 & 25.0 & 21.0 & 29.0 \\
        $\pi_0$ \textbf{+ TACO (RND)} & 54.0 & 33.0 & \textbf{30.0} & 39.0\\
        $\pi_0$ \textbf{+ TACO (CFN, Ours)} & \textbf{62.0} & \textbf{40.0} & 27.0 & \textbf{43.0} \\
            
        \bottomrule
        \end{tabular}
    }
\end{table}

\section{Additional Real-World Experiment Details}

\begin{figure*}[t]
    \centering
    \resizebox{0.9\textwidth}{!}{%
        \includegraphics{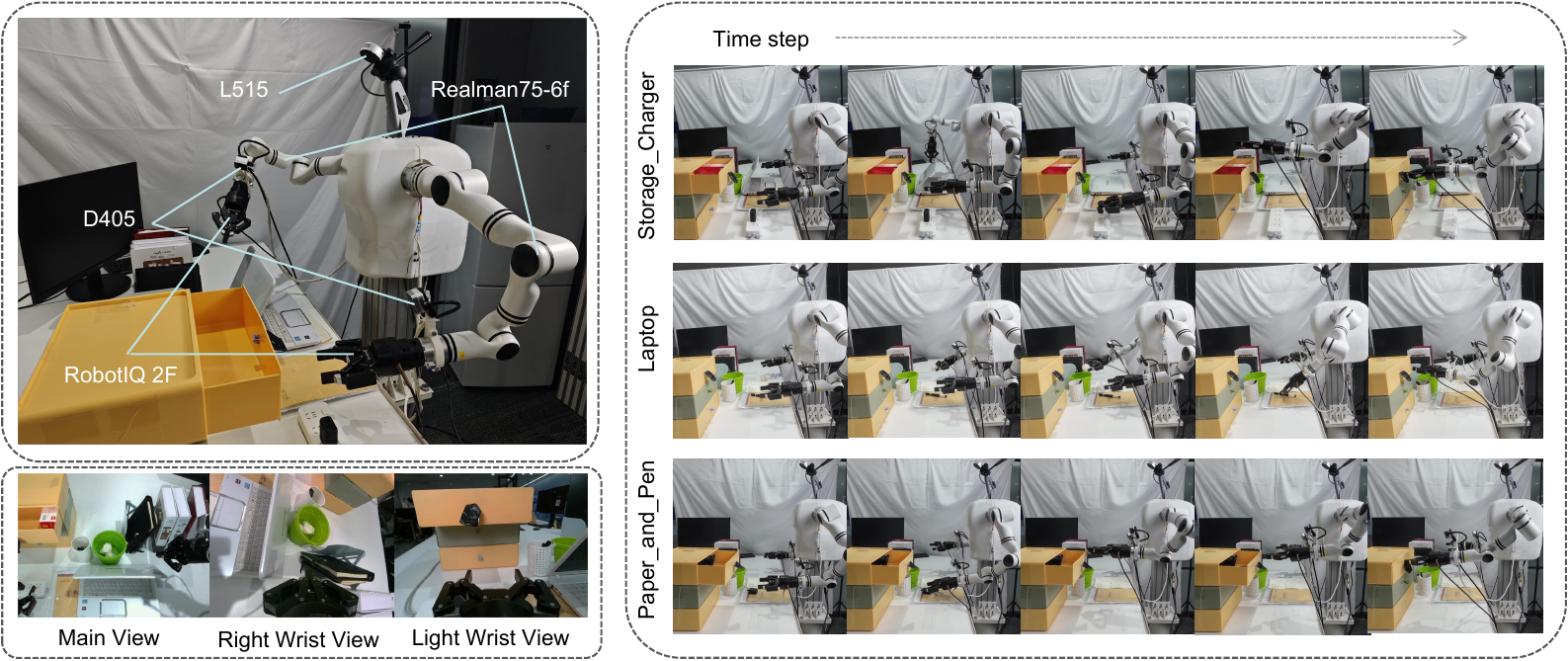}
    }
    \caption{Schematic illustration of our real-world robotic platform (left), and the task execution pipeline for the three real-world tasks (right).}
    \label{fig:real_setting}
\end{figure*}

\textbf{Real-World Platform and Task Setup.}
As shown in Figure \ref{fig:real_setting}, 
our platform consists of a Realman RM75-6F dual-arm robot, two Robotiq 2F grippers, one Intel RealSense L515 camera serving as the main-view camera, and two Intel RealSense D405 cameras mounted on the wrists.
including one main-view camera and two wrist-mounted cameras. 
A workstation equipped with an RTX 4090 GPU is used for model deployment and inference. 

The prompts for each task are listed in Table \ref{tab:prompt}. 
During testing, each object is randomly placed within an area of approximately $3\,\text{cm} \times 4\,\text{cm}$. 
The overall task execution process is illustrated in Figure \ref{fig:real_setting}.

\begin{table*}[t]
\centering
\begin{tabular}{p{3.2cm} p{10cm}}
\hline
\textbf{Task} & \textbf{Prompt} \\
\hline

Receive Book & Use the right arm to catch the book handed over by the person. \\

Paper and Pen & Place the paper ball into the trash bin with the left arm, and put the pen into the pen holder with the right arm. \\

Storage Charger & The left arm places the charging head into the drawer and closes it. \\

Pick Books & Grip the book with both hands, one hand on each side, and lift it simultaneously with both arms. \\

Laptop & Close the laptop lid using the right arm, unplug the charger with the left arm, place it in the top drawer, and close the drawer. \\

\hline
\end{tabular}
\caption{Tasks in real world and their corresponding prompts.}
\label{tab:prompt}
\end{table*}

\end{document}